\documentclass[conference]{IEEEtran}
\IEEEoverridecommandlockouts
\usepackage{tabularx}
\usepackage{cite}
\usepackage{amsmath,amssymb,amsfonts}
\usepackage{graphicx}
\usepackage{textcomp}
\usepackage{xcolor}
\usepackage[font=small,labelfont=bf]{caption}
\usepackage{graphicx,xspace,verbatim,comment}
\usepackage{hyperref,array,color,balance,multirow}
\usepackage{balance,float,url,amsfonts,alltt}
\usepackage{mathtools,rotating,amsmath,amssymb}
\usepackage{color,ifpdf,fancyvrb}
\usepackage{etoolbox,listings,subcaption}
\usepackage{bigstrut,morefloats,pbox}
\usepackage{amsmath}
\usepackage{algorithm}
\usepackage[noend]{algpseudocode}
\usepackage{booktabs}
\usepackage{bm}
\def\BibTeX{{\rm B\kern-.05em{\sc i\kern-.025em b}\kern-.08em
    T\kern-.1667em\lower.7ex\hbox{E}\kern-.125emX}}

\title{Predicting Eating Events in Free Living Individuals - A Technical Report
\thanks{This research was supported by grant U54EB020404 awarded by the National Institute of Biomedical Imaging and Bioengineering and by grant 5R01CA164993 awarded by the National Cancer Institute.}
}

\makeatletter
\newcommand{\linebreakand}{%
	\end{@IEEEauthorhalign}
	\hfill\mbox{}\par
	\mbox{}\hfill\begin{@IEEEauthorhalign}
}
\makeatother

\author{
 \IEEEauthorblockN{Jiayi Wang}
    \IEEEauthorblockA{\textit{Computer Science and Engineering} \\
        \textit{University of California San Diego}\\
        San Diego, USA \\
        jiw433@ucsd.edu}
    \and
    \IEEEauthorblockN{Jiue-An Yang}
    \IEEEauthorblockA{\textit{Qualcomm Institute / Calit2} \\
        \textit{University of California San Diego}\\
        San Diego, USA \\
        jayyang@eng.ucsd.edu}
    \and
  
    \IEEEauthorblockN{Supun Nakandala}
    \IEEEauthorblockA{\textit{Computer Science and Engineering} \\
        \textit{University of California San Diego}\\
        San Diego, USA \\
        snakanda@eng.ucsd.edu}
    \linebreakand
    \IEEEauthorblockN{Arun Kumar}
    \IEEEauthorblockA{\textit{Computer Science and Engineering} \\
        \textit{University of California San Diego}\\
        San Diego, USA \\
        arunkk@eng.ucsd.edu}
    \and
    \IEEEauthorblockN{Marta M. Jankowska}
    \IEEEauthorblockA{\textit{Qualcomm Institute / Calit2} \\
        \textit{University of California San Diego}\\
        San Diego, USA \\
        majankowska@ucsd.edu}
}

\begin{document}

\maketitle

\begin{abstract}
This technical report records the experiments of applying multiple machine learning algorithms for predicting eating and food purchasing behaviors of free-living individuals. Data was collected with accelerometer, global positioning system (GPS), and body-worn cameras called SenseCam over a one week period in 81 individuals from a variety of ages and demographic backgrounds. These data were turned into minute-level features from sensors as well as engineered features that included time (e.g., time since last eating) and environmental context (e.g., distance to nearest grocery store). Algorithms include Logistic Regression, RBF-SVM, Random Forest, and Gradient Boosting. Our results show that the Gradient Boosting model has the highest mean accuracy score (0.7289) for predicting eating events before 0 to 4 minutes. For predicting food purchasing events, the RBF-SVM model (0.7395) outperforms others. For both prediction models, temporal and spatial features were important contributors to predicting eating and food purchasing events.
\end{abstract}

\section{Introduction}
In this pilot study, we explore the application of multiple machine learning models on predicting eating and food purchasing events from body worn sensor data as well as contextual environment data. We utilize a proof of concept sample of 81 free-living individuals with significant sensor data. We detail the training data and feature generation for model input.  We explore the ability of the models to predict an outcome "in the moment", as well as using a time offset to assess their utility in predicting eating/food purchasing events up to 4 minutes in advance. We present the results for two binary classification problems: 1) predict eating events, and 2) predict food purchasing events. Both problems received five tasks with the same features, and each task receiving a different time offset (0-4 minutes) of the data (e.g., task 1 time point of features matched time point of event, task 2 time point of features were 1 minute prior to the event, and so on). For each task, we present 4 classifiers, including Logistic Regression, RBF-SVM, Random Forest, Gradient Boosting.  This technical report details the data, feature engineering, model construction, and model evaluation for the study. Background of the research project and discussion of the study results are described in an upcoming proceeding paper [reference will be updated upon availability].  

\section{Data}
\subsection{Study Sample}
A sub-sample of 81 participants were selected from an observational research cohort conducted in 2012-2015 of 216 individuals living in San Diego County between the ages of 6 and 85 years.  Participants in the study wore multiple sensor devices including a SenseCam camera, a hip-worn GPS device (Qstarz BT-Q1000XT), and a hip-worn triaxial accelerometer (GT3X+, ActiGraph). Participants were asked to wear the devices for 7 days during waking hours and were asked to re-wear devices if certain wear time criteria (a minimum of 600 minutes per day) were not met. Approval for this study is under UCSD IRB protocol 111160. 

\subsection{Training Data}
\begin{figure}[!htb]
\centering
   \begin{subfigure}{0.493\linewidth}
   	\includegraphics[width=\linewidth]{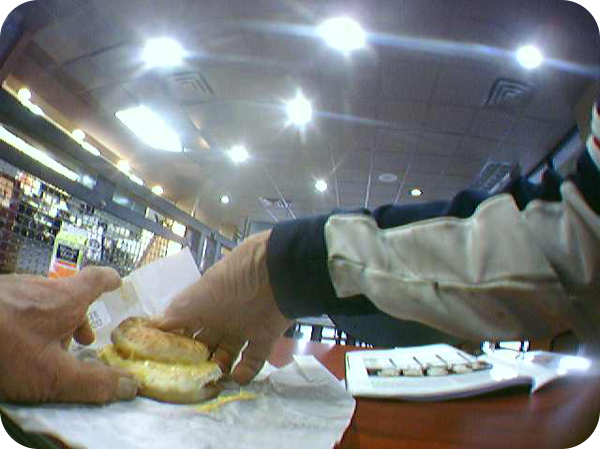}
   \end{subfigure}
   \begin{subfigure}{0.493\linewidth}
   	\includegraphics[width=\linewidth]{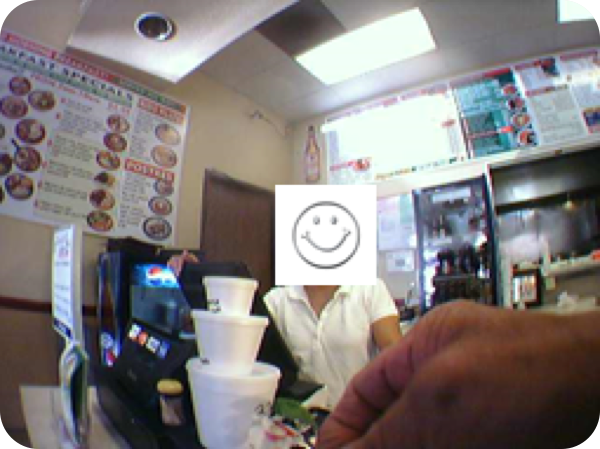}
   \end{subfigure}
\caption{Examples of tagged eating event (left) and food purchasing (right) from the front-facing SenseCam imagery data.}
\label{fig:SenseCame_IMG}
\end{figure}

Training data for this study was generated from the SenseCam images, which provide objective records of eating and food purchasing behavior. At the minute level, eating and food purchasing events were tagged by a team of image coders trained in a detailed protocol with quality control for accurate coding performed on a subset of images \cite{Chen2013}. Examples of eating and food purchasing images are shown in Figure \ref{fig:SenseCame_IMG}.

\subsection{Sensor Data}
GPS and accelerometer data was processed with the Personal Activity Location Measurement System (PALMS), which filters spurious data points, smooths out common GPS interference patterns, and aggregates data to a one minute epoch. For each data point, three GPS features were extracted including distance from last point in meters, speed in km/h, and change in direction between points. Additionally, eight acceleration features were extracted including magnitude of the 3-axis of acceleration, vector magnitude, activity intensity (counts per minute of the accelerometer), and ambient light. PALMS also generates features from combined GPS and accelerometer inputs including if the participant was likely to be wearing or not wearing the devices, if they were outside/inside/in a vehicle, if they were stationary/walking/bicycling/driving, and the day of the week. Minutes with no GPS data, or with a location beyond the San Diego County boundaries were removed from analysis. 

\subsection{Engineered Features}
Several features were engineered using sensor and contextual data about the environment pertinent to eating and food purchasing behaviors (Table \ref{tbl:extracted_features}). Home locations of the participants were identified by spatially clustering the sleep time data points (3:00-4:00am) over the study period with DBSCAN, which is then used to create the \textit{In Home} feature with a 50-meter home radius. In order to captured the distance proximity to retail outlets, business locations were extracted from the 2016 Esri Business Analyst data. Distances to outlets of five NAICS categories (445 - Food and Beverage Store, 446 - Health and Personal Care Store, 447 - Gasoline Station, 7224 - Drinking Place, and 7225 - Full Service Restaurant) were used to represent the contextual retail food environments of the participants. Time features were developed to both account for time of day, as well as time since last activity of interest including bouts of physical activity, eating, and food purchasing.

\begin{table}[ht!]
\centering
\caption{Engineered Features}
\label{tbl:extracted_features}
\begin{tabular}{m{2cm} | m{5cm} | m{0.5cm}}
\toprule
	Name & Description & Size \\
\midrule
Time & One hot encoding form of 6 time ranges within the day (0:00-6:00,  6:00-10:00, 10:00-14:00, 14:00-17:00, 17:00-20:00, and 20:00-23:59). & 6\\
\midrule
In Home & Participant is in their home defined as being within 50 meters of inferred home location. & 1 \\
\midrule
Time Since Activity Bout & Number of minutes from last sedentary, physical activity, or moderate to vigorous physical activity bout to current time (value of data under activity bout event is set to 0). & 3\\
\midrule
Time Since Eating & Number of minutes from last eating event. & 1\\
\midrule
Time Since Purchasing & Number of minutes from last purchasing event. & 1\\
\midrule
Distance to Retail Outlet & Participant's distance in meters to the nearest retail outlet in five categories: grocery stores (e.g., grocery, meat/fish, dairy products, etc.), health and personal care stores (e.g., drug stores, medical equipment, miscellaneous retail stores, etc.), gasoline stations (with or without convenience stores), drinking places (e.g., bar, brewery, nightclub), and restaurants and other eating places. & 5\\ 
\midrule
Time Pattern & The distance from current time to middle point in corresponding time intervals (6am-9am, 11am-14pm, 17pm-20pm) For example, 8am falls into 6am-9am with the middle point of 7:30am, thus its value is 30. & 1\\
\bottomrule
\end{tabular}
\end{table}

\section{Model Description}
\subsection{Dataset}
Data from the 81 individuals contains a total of 596,771 data points, and each data point includes features from accelerometer, GPS, and engineered time sensitive and context features (Table \ref{tbl:extracted_features}) at the minute level. We select 60 individuals as the training set (428,874 data points) and the rest as the test set (167,897 data points). The tasks of predicting eating and tasks of predicting food purchasing share the same dataset and features but have different positive or negative labels for each data point based on the corresponding behavior. A total of 34 sensor and engineered features are used for the Logistic Regression model while other models (RBF-SVM, Random Forest, and Gradient Boosting) used 35 features. The one different feature is the numerical time feature (e.g. 14:30pm is coded as 14.5).  Different from the other three models that handle non-linear features, Logistic Regression is a linear model thus the numerical time features were excluded and the time information is covered by our engineered 6 variables representing the time intervals.

\subsection{Methodology}
When training with 428,874 data points of 60 individuals, we performed 5-fold nested cross-validation for fine tuning hyper-parameters, where we randomly selected 15 out of 60 individuals as the validation set and the remaining 45 individuals as the training set in each fold. Because each individual gave different numbers of data points, five folds in cross validation have different sizes of training and validation sets. In addition, because the ratio of eating time or food purchasing time to the time of the rest of a day is very small, the dataset is highly imbalanced. To alleviate the imbalance problem, we sample the negative data pool to have the same amount of negative data as positive data. For tasks of predicting eating, five folds have 10,972, 10,524, 11,782, 11,174, and 9,950 training data respectively after sampling; for tasks of predicting food purchasing, five folds have 484, 378, 482, 488, and 428 training data after sampling; and five folds of both types of tasks have 115,587, 115,224, 99,405, 101,575, and 123,870 validation data respectively. We use all four classifiers with the scikit-learn package\cite{sklearn_api}\cite{scikit-learn}.

\vspace{5mm}

\begingroup
\raggedright
\paragraph{Logistic Regression} For the LR model we use \textit{sklearn.linear.model.LogisticRegression} with L2 regularization, and fine tune the hyper-parameter C which is the inverse of regularization strength. The grid size is as follows: C $\in \{1000, 100, 10, 1, 0.1\}$.\\
\vspace{5mm}
\raggedright
\paragraph{RBF-SVM} For the RBF-SVM model we use \textit{sklearn.svm.SVC} with arguments kernel = 'rbf'. For tasks of predicting food purchasing, we use max$\_$iter = 10000 in both cross validation and final training. For tasks of predicting eating, we use max$\_$iter = 1000 for cross validation and max$\_$iter = 10000 for final training. We fine tune two hyper-parameters, $\gamma$ and C. $\gamma$ controls the bandwidth in the Gaussian kernel, and C is the cost of misclassification. The grid size is as follows:  $\gamma \in \{0.01, 0.1, 1, 10, 100\}$, and C $\in \{1000, 100, 10, 1, 0.1\}$.\\
\vspace{5mm}
\raggedright
\paragraph{Random Forest} For the RF model we use \textit{sklearn.ensemble.RandomForestClassifier}. We fine tune two hyper-parameters, n{\_}estimators and max{\_}depth. n{\_}estimators controls the number of trees in RF, and max{\_}depth controls the maximum height of the trees. The grid size is as follows: n{\_}estimators $\in \{10, 30, 50, 100, 200\}$, and max{\_}depth $\in \{2, 5, 10, 15, 20\}$.\\
\vspace{5mm}
\raggedright
\paragraph{Gradient Boosting} For the GB model we use \textit{sklearn.ensemble.GradientBoostingClassifier}. We fine tune the same two hyper-parameters, n{\_}estimators and max{\_}depth. The grid size is as follows: n{\_}estimators $\in$ $\{10, 30, 50, 100, 200\}$, and max{\_}depth $\in$ $\{1, 2, 3, 4, 5\}$.\\
\endgroup

\section{Results}
\begin{figure*}
\centering
  \begin{subfigure}{0.493\linewidth}
  	\includegraphics[height=6cm, width=\linewidth]{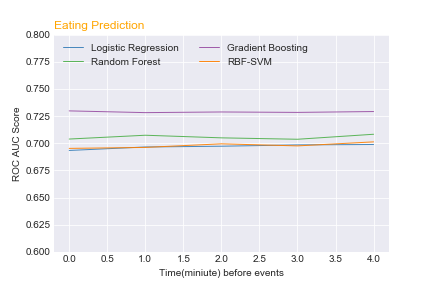}
  \end{subfigure}
  \begin{subfigure}{0.493\linewidth}
  	\includegraphics[height=6cm, width=\linewidth]{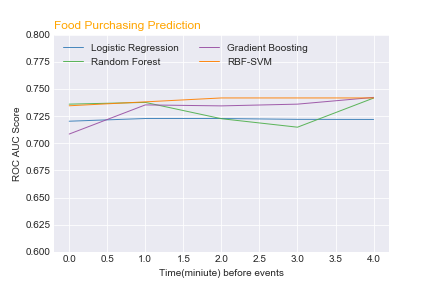}
  \end{subfigure}
\caption{Accuracy results for predicting predicting eating (left) and purchasing (right) events over 5 timepoints (0 - 4 minutes) for all four models.}
\label{fig:ROC_AUC_curve_IMG}
\end{figure*}

We use the ROC AUC score as our metric from scikit-learn to evaluate the performance of the models. For binary classification problems, ROC AUC score is the same as balanced accuracy, which is the average of recall of both classes. Because our dataset is imbalanced, balanced accuracy is preferred. We fine-tuned the hyper-parameters by using 5-fold cross-validation (Table \ref{tbl:Logistic Regression_Eating_cv_results_min0} to Table \ref{tbl:Gradient Boosting_Food Purchasing_cv_results_min4}) on four models, and we evaluate performance of Logistic Regression (Table \ref{tbl:lr_results}), RBF-SVM (Table \ref{tbl:svm_results}), Random Forest (Table \ref{tbl:rf_results}) and Gradient Boosting (Table \ref{tbl:gb_results}) on tasks of predicting eating and tasks of predicting food purchasing. Based on the ROC AUC scores (Figure \ref{fig:ROC_AUC_curve_IMG}), Gradient Boosting outperformed others for predicting eating with a mean accuracy of 0.7289 of 0-4 mins and the RBF-SVM model stood out in predicting food purchasing with a mean accuracy of 0.7395. For tasks of predicting eating, models are not sensitive to time offsets as the curves of models are flat across time offsets. For tasks of predicting food purchasing, RBF-SVM and Logistic Regression models are flat, and Random Forest and Gradient Boosting models fluctuate. Model performances by the order of ROC AUC scores(Table \ref{tbl:mean_acc_four_models}) are as follows:  

\vspace{3mm}
\begin{itemize}
  \item Model performance for predicting \textbf{\textit{eating}} events: Gradient Boosting $>$ Random Forest $>$ RBF-SVM $>$ Logistic Regression
  \item Model performance for predicting \textbf{\textit{food purchasing }} events: RBF-SVM $>$ Gradient Boosting $>$ Random Forest $>$ Logistic Regression
\end{itemize}
\vspace{3mm}

\begin{figure*}
\centering
  \begin{subfigure}{0.493\linewidth}
  	\includegraphics[width=\linewidth]{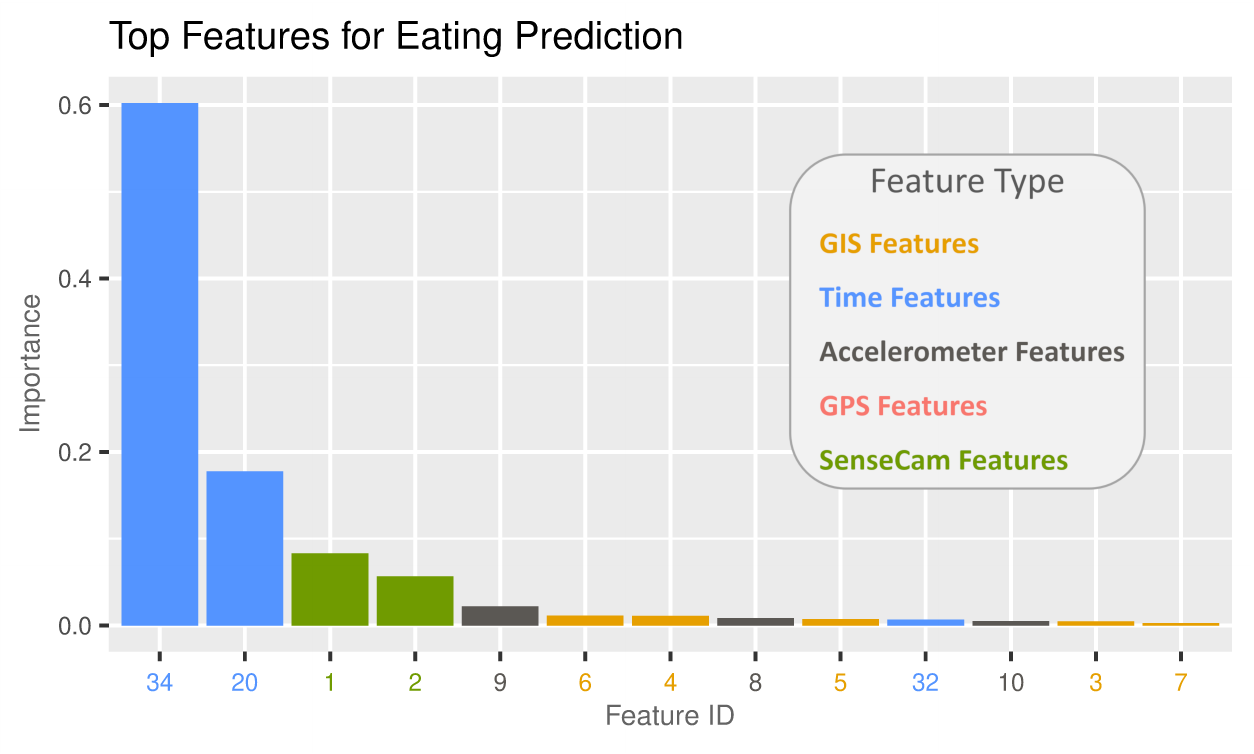}
  \end{subfigure}
  \begin{subfigure}{0.493\linewidth}
  	\includegraphics[width=\linewidth]{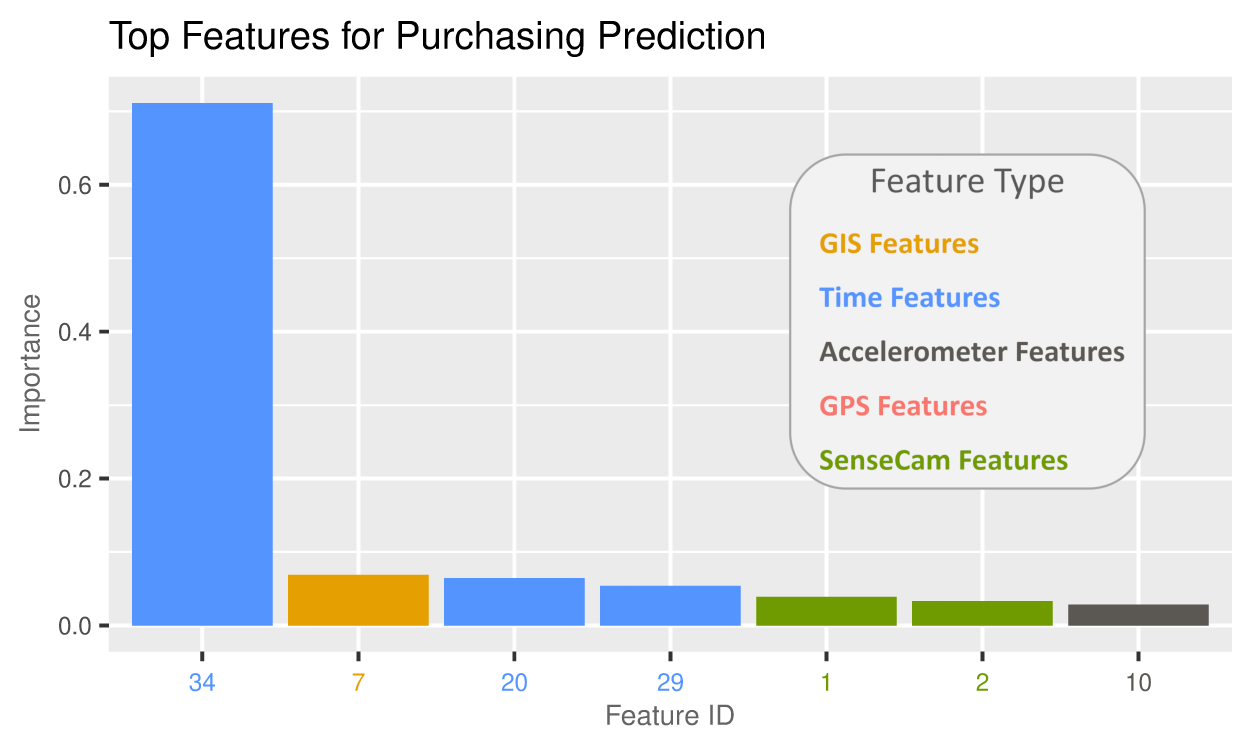}
  \end{subfigure}
\caption{Top features for predicting eating (left) and purchasing (right) events. Colors represent the type of features.}
\label{featureImportance_IMG}
\end{figure*}

The top features of the Gradient Boosting model for predicting both events in the order of feature importance is reported in Figure \ref{featureImportance_IMG}. Some top features that contributed to both event predictions included:
\begin{itemize}
  \item \textbf{temporal} features: \textit{time (34)}, \textit{time pattern(20)}, \textit{time since last eating (1)}, \textit{time since last food purchasing (2)}.
  \item \textbf{spatial} features: \textit{distance from nearest health care places (4)}, \textit{distance from nearest Gasoline Station (5)}, \textit{distance from nearest food and beverage stores (3)}.
\end{itemize}

\bibliography{techReportVersion2}

\newcommand{\noop}[1]{}
\begin{thebibliography}{1}

\bibitem{sklearn_api}
Lars Buitinck, Gilles Louppe, Mathieu Blondel, Fabian Pedregosa, Andreas
  Mueller, Olivier Grisel, Vlad Niculae, Peter Prettenhofer, Alexandre
  Gramfort, Jaques Grobler, Robert Layton, Jake VanderPlas, Arnaud Joly, Brian
  Holt, and Ga{\"{e}}l Varoquaux.
\newblock {API} design for machine learning software: experiences from the
  scikit-learn project.
\newblock In {\em ECML PKDD Workshop: Languages for Data Mining and Machine
  Learning}, pages 108--122, 2013.

\bibitem{Chen2013}
Jacqueline Chen, Simon~J. Marshall, Lu~Wang, Suneeta Godbole, Amanda Legge,
  Aiden Doherty, Paul Kelly, and et~al.
\newblock {Using the SenseCam as an objective tool for evaluating eating
  patterns}.
\newblock In {\em Proceedings of the 4th International SenseCam {\&} Pervasive
  Imaging Conference on - SenseCam '13}, pages 34--41, 2013.
\newblock URL: \url{https://dl.acm.org/citation.cfm?id=2526673}, \href
  {https://doi.org/10.1145/2526667.2526673}
  {\path{doi:10.1145/2526667.2526673}}.

\bibitem{scikit-learn}
F.~Pedregosa, G.~Varoquaux, A.~Gramfort, V.~Michel, B.~Thirion, O.~Grisel,
  M.~Blondel, P.~Prettenhofer, R.~Weiss, V.~Dubourg, J.~Vanderplas, A.~Passos,
  D.~Cournapeau, M.~Bruocher, M.~Perrot, and E.~Duchesnay.
\newblock Scikit-learn: Machine learning in {P}ython.
\newblock {\em Journal of Machine Learning Research}, 12:2825--2830, 2011.

\end{thebibliography}

\begin{table}[ht!]
\centering
\small
\caption{Feature ID and Name}
\label{tbl:all_features}
\begin{tabular}{m{2cm} | m{5cm}}
\toprule
	Feature ID & Feature Name\\
\midrule
0&bias\\
\midrule
1&time since last eating\\
\midrule
2&time since last food purchasing\\
\midrule
3&distance from nearest FoodBeverage stores\\
\midrule
4&distance from nearest Health Care places\\
\midrule
5&distance from nearest Gasolin Station\\
\midrule
6&distance from nearest Drinks places\\
\midrule
7&distance from nearest Eating places\\
\midrule
8&time since last stationary activity\\
\midrule
9&time since last physical activity\\
\midrule
10&time since last moderate or rigorous activity\\
\midrule
11&activity\\
\midrule
12&axis2\\
\midrule
13&axis3\\
\midrule
14&distance\\
\midrule
15&speed\\
\midrule
16&vectorMag\\
\midrule
17&lux\\
\midrule
18&wearing\\
\midrule
19&eat at home\\
\midrule
20&time pattern\\
\midrule
21&Monday\\
\midrule
22&Tuesday\\
\midrule
23&Wednesday\\
\midrule
24&Thursday\\
\midrule
25&Friday\\
\midrule
26&Saturday\\
\midrule
27&Sunday\\
\midrule
28&0am-6am\\
\midrule
29&6am-10am\\
\midrule
30&10am-14pm\\
\midrule
31&14pm-17pm\\
\midrule
32&17pm-20pm\\
\midrule
33&20pm-23:59pm \\
\midrule
34&time\\

\bottomrule
\end{tabular}
\end{table}

\begin{table}[ht!]
\centering
\caption{Mean Accuracy of 0-4 Minutes}
\label{tbl:mean_acc_four_models}
\begin{tabular}{p{1.3cm} p{0.6cm} p{0.6cm} p{0.6cm}| p{0.6cm}  p{0.6cm} p{0.6cm}}
\toprule
\multicolumn{4}{c|}{Eating} & \multicolumn{3}{|c}{Food Purchasing}\\
\midrule
Models & Train Acc. & Valid. Acc. & Test Acc. & Train Acc. & Valid. Acc. & Test Acc. \\
\midrule
LR&0.6934&0.6748&0.6969& 0.739 &0.6983& 0.722\\
\midrule
RBF-SVM&0.6952&0.6107&0.698& 0.7386&0.7013&\textbf{0.7395}\\
\midrule
RF&0.733&0.6875&0.7057& 0.8067 &0.671& 0.7305\\
\midrule
GB&0.7356&0.7039&\textbf{0.7289}& 0.775 &0.6983& 0.7313\\
\bottomrule
\end{tabular}
\end{table}

\begin{table}[ht!]
\centering
\caption{Logistic Regression Model Results}
\label{tbl:lr_results}
\begin{tabular}{p{0.25cm} p{0.6cm} p{0.6cm} p{0.6cm} p{0.25cm} | p{0.6cm} p{0.6cm} p{0.6cm} p{0.25cm}}

\toprule
\multicolumn{5}{c|}{Eating} & \multicolumn{4}{|c}{Food Purchasing}\\
\midrule
Mins & Train Acc. & Valid. Acc. & Test Acc. & C & Train Acc. & Valid. Acc. & Test Acc. & C\\
\midrule
0&0.6928&0.6736&0.6934&0.1&0.7417&0.6977&0.7203&0.1\\
\midrule
1&0.6933&0.6742&0.6966&0.1&0.7350&0.7136&\textbf{0.7228}&0.1\\
\midrule
2&0.6932&0.6745&0.6973&0.1&0.7383&0.6904&\textbf{0.7228}&0.1\\
\midrule
3&0.6937&0.6757&0.6985&0.1&0.7400&0.6956&0.7220&0.1\\
\midrule
4&0.6939&0.6759&\textbf{0.6989}&0.1&0.7400&0.6944&0.7219&0.1\\
\bottomrule
\end{tabular}
\end{table}

\begin{table}[ht!]
\centering
\caption{RBF-SVM Model Results}
\label{tbl:svm_results}
\begin{tabular}{p{0.25cm} p{0.5cm} p{0.5cm} p{0.5cm} p{0.5cm} p{0.5cm}| p{0.5cm} p{0.5cm} p{0.5cm} p{0.5cm} p{0.5cm}}
\toprule
\multicolumn{6}{c|}{Eating} & \multicolumn{5}{|c}{Food Purchasing}\\
\midrule
Mins & Train Acc. & Valid. Acc. & Test Acc. & $\gamma$ & C & Train Acc. & Valid. Acc. & Test Acc. &$\gamma$ &C \\
\midrule
0&0.6924&0.6203&0.6953&0.1&0.1&0.7400&0.7055&0.7345&0.1&1\\
\midrule
1&0.6930&0.6040&0.6962&0.01&10&0.7383&0.7053&0.7381&0.1&1\\
\midrule
2&0.6980&0.6126&0.6995&0.01&100&0.7383&0.6987&\textbf{0.7417}&0.01&10\\
\midrule
3&0.6940&0.6076&0.6976&0.01&1&0.7383&0.6986&\textbf{0.7417}&0.1&1\\
\midrule
4&0.6988&0.6088&\textbf{0.7013}&0.01&100&0.7383&0.6985&\textbf{0.7417}&0.1&1\\
\bottomrule
\end{tabular}
\end{table}

\begin{table}[ht!]
\centering
\caption{Random Forest Model Results}
\label{tbl:rf_results}
\begin{tabular}{p{0.25cm} p{0.5cm} p{0.5cm} p{0.5cm} p{0.5cm} p{0.5cm}| p{0.5cm} p{0.5cm} p{0.5cm} p{0.5cm} p{0.5cm}}
\toprule
\multicolumn{6}{c|}{Eating} & \multicolumn{5}{|c}{Food Purchasing}\\
\midrule
Mins & Train Acc. & Valid. Acc. & Test Acc. & Num. Tree & Max Depth & Train Acc. & Valid. Acc. & Test Acc. & Num. Tree & Max Depth\\
\midrule
0&0.7227&0.6897&0.7039&10&5&0.7717&0.6740&0.7360&200&2\\
\midrule
1&0.7374&0.6878&0.7074&200&5&0.7667&0.6643&0.7376&200&2\\
\midrule
2&0.7281&0.6877&0.7050&10&5&0.8800&0.6693&0.7226&200&5\\
\midrule
3&0.7377&0.6857&0.7037&200&5&0.8500&0.6706&0.7148&10&5\\
\midrule
4&0.7391&0.6867&\textbf{0.7083}&200&5&0.7650&0.6769&\textbf{0.7416}&200&2\\
\bottomrule
\end{tabular}
\end{table}

\begin{table}[ht!]
\centering
\caption{Gradient Boosting Model Results}
\label{tbl:gb_results}
\begin{tabular}{p{0.25cm} p{0.5cm} p{0.5cm} p{0.5cm} p{0.5cm} p{0.5cm}| p{0.5cm} p{0.5cm} p{0.5cm} p{0.5cm} p{0.5cm}}
\toprule
\multicolumn{6}{c|}{Eating} & \multicolumn{5}{|c}{Food Purchasing}\\
\midrule
Mins & Train Acc. & Valid. Acc. & Test Acc. & Num. Tree & Max Depth & Train Acc. & Valid. Acc. & Test Acc. & Num. Tree & Max Depth\\
\midrule
0&0.7335&0.7073&\textbf{0.7298}&200&1&0.7783&0.6977&0.7085&10&2\\
\midrule
1&0.7382&0.7056&0.7282&200&1&0.7800&0.7136&0.7353&10&2\\
\midrule
2&0.7370&0.7042&0.7288&200&1&0.7733&0.6904&0.7344&10&2\\
\midrule
3&0.7354&0.7018&0.7284&200&1&0.7633&0.6956&0.7360&10&2\\
\midrule
4&0.7339&0.7007&0.7292&200&1&0.7800&0.6944&\textbf{0.7422}&10&2\\
\bottomrule
\end{tabular}
\end{table}

\onecolumn

\newpage
\onecolumn

\begin{table}[ht!]
\centering
\caption{5-fold Cross Validation Logistic Regression Model Results: 0 Min Before Eating}
\label{tbl:Logistic Regression_Eating_cv_results_min0}
\begin{tabular}{|p{1cm}| p{1cm} |p{1cm}| p{1cm}| p{1cm}| p{1cm}| p{1cm}| p{1cm} |p{1cm} |p{1cm} |p{1cm}| p{1cm} |p{1cm}|}
\toprule
\multirow{2}{*}{C} &\multicolumn{2}{c|}{fold1} & \multicolumn{2}{c|}{fold2} & \multicolumn{2}{c|}{fold3} & \multicolumn{2}{c|}{fold4} & \multicolumn{2}{c|}{fold5}  & \multicolumn{2}{c|}{average} \\\cline{2-13}
&\multicolumn{1}{c|}{Train} & \multicolumn{1}{c|}{Valid.} & \multicolumn{1}{c|}{Train} & \multicolumn{1}{c|}{Valid.} & \multicolumn{1}{c|}{Train}  & \multicolumn{1}{c|}{Valid.} &\multicolumn{1}{c|}{Train} & \multicolumn{1}{c|}{Valid.} & \multicolumn{1}{c|}{Train} & \multicolumn{1}{c|}{Valid.} & \multicolumn{1}{c|}{Train}  & \multicolumn{1}{c|}{Valid.}\\
\midrule
1000&0.7021&\textbf{0.6548}&0.6951&0.652&0.6799&0.6944&0.6862&0.664&0.6921&0.6659&0.6911&0.6662\\
\midrule
100&0.7024&0.6535&0.695&0.6549&0.6793&0.695&0.6859&0.6674&0.6921&0.6654&0.6909&0.6673\\
\midrule
10&0.7026&0.647&0.6933&0.6651&0.679&0.6959&0.6872&0.6828&0.6923&0.6638&0.6909&0.6709\\
\midrule
1&0.7035&0.6444&0.6899&\textbf{0.6664}&0.6755&0.6972&0.6786&0.6828&0.6936&\textbf{0.6674}&0.6882&0.6716\\
\midrule
0.1&0.6991&0.6486&0.6883&0.6644&0.6737&\textbf{0.7033}&0.6787&\textbf{0.6864}&0.6903&0.665&0.686&\textbf{0.6736}\\
\bottomrule
\end{tabular}
\end{table}

\begin{table}[ht!]
\centering
\caption{5-fold Cross Validation Logistic Regression Model Results: 1 Min Before Eating}
\label{tbl:Logistic Regression_Eating_cv_results_min1}
\begin{tabular}{|p{1cm}| p{1cm} |p{1cm}| p{1cm}| p{1cm}| p{1cm}| p{1cm}| p{1cm} |p{1cm} |p{1cm} |p{1cm}| p{1cm} |p{1cm}|}
\toprule
\multirow{2}{*}{C} &\multicolumn{2}{c|}{fold1} & \multicolumn{2}{c|}{fold2} & \multicolumn{2}{c|}{fold3} & \multicolumn{2}{c|}{fold4} & \multicolumn{2}{c|}{fold5}  & \multicolumn{2}{c|}{average} \\\cline{2-13}
&\multicolumn{1}{c|}{Train} & \multicolumn{1}{c|}{Valid.} & \multicolumn{1}{c|}{Train} & \multicolumn{1}{c|}{Valid.} & \multicolumn{1}{c|}{Train}  & \multicolumn{1}{c|}{Valid.} &\multicolumn{1}{c|}{Train} & \multicolumn{1}{c|}{Valid.} & \multicolumn{1}{c|}{Train} & \multicolumn{1}{c|}{Valid.} & \multicolumn{1}{c|}{Train}  & \multicolumn{1}{c|}{Valid.}\\
\midrule
1000&0.7022&\textbf{0.6553}&0.6956&0.6541&0.6801&0.6964&0.6874&0.666&0.6935&0.6652&0.6917&0.6674\\
\midrule
100&0.7012&0.654&0.6953&0.6573&0.6799&0.697&0.6877&0.6683&0.693&0.6658&0.6914&0.6685\\
\midrule
10&0.703&0.6457&0.6932&0.6653&0.68&0.6979&0.687&0.6833&0.693&\textbf{0.6667}&0.6912&0.6718\\
\midrule
1&0.7044&0.6429&0.691&\textbf{0.6653}&0.6756&0.7001&0.6794&0.6854&0.6936&0.6649&0.6888&0.6717\\
\midrule
0.1&0.7&0.6478&0.689&0.665&0.6737&\textbf{0.7056}&0.6798&\textbf{0.6872}&0.6903&0.6652&0.6865&\textbf{0.6742}\\
\bottomrule
\end{tabular}
\end{table}

\begin{table}[ht!]
\centering
\caption{5-fold Cross Validation Logistic Regression Model Results: 2 Mins Before Eating}
\label{tbl:Logistic Regression_Eating_cv_results_min2}
\begin{tabular}{|p{1cm}| p{1cm} |p{1cm}| p{1cm}| p{1cm}| p{1cm}| p{1cm}| p{1cm} |p{1cm} |p{1cm} |p{1cm}| p{1cm} |p{1cm}|}
\toprule
\multirow{2}{*}{C} &\multicolumn{2}{c|}{fold1} & \multicolumn{2}{c|}{fold2} & \multicolumn{2}{c|}{fold3} & \multicolumn{2}{c|}{fold4} & \multicolumn{2}{c|}{fold5}  & \multicolumn{2}{c|}{average} \\\cline{2-13}
&\multicolumn{1}{c|}{Train} & \multicolumn{1}{c|}{Valid.} & \multicolumn{1}{c|}{Train} & \multicolumn{1}{c|}{Valid.} & \multicolumn{1}{c|}{Train}  & \multicolumn{1}{c|}{Valid.} &\multicolumn{1}{c|}{Train} & \multicolumn{1}{c|}{Valid.} & \multicolumn{1}{c|}{Train} & \multicolumn{1}{c|}{Valid.} & \multicolumn{1}{c|}{Train}  & \multicolumn{1}{c|}{Valid.}\\
\midrule
1000&0.7009&\textbf{0.6544}&0.6969&0.6561&0.6809&0.6977&0.6871&0.6719&0.6942&\textbf{0.6664}&0.692&0.6693\\
\midrule
100&0.7014&0.652&0.6962&0.6588&0.6812&0.698&0.6875&0.6719&0.6938&0.6657&0.692&0.6693\\
\midrule
10&0.7043&0.6446&0.6924&\textbf{0.6655}&0.6806&0.6999&0.6864&0.684&0.6943&0.6663&0.6916&0.6721\\
\midrule
1&0.7049&0.6411&0.6919&0.6649&0.6755&0.701&0.6795&0.6891&0.6952&0.6658&0.6894&0.6724\\
\midrule
0.1&0.7006&0.6474&0.6899&0.6653&0.6733&\textbf{0.7053}&0.6802&\textbf{0.6891}&0.6899&0.6652&0.6868&\textbf{0.6745}\\
\bottomrule
\end{tabular}
\end{table}

\begin{table}[ht!]
\centering
\caption{5-fold Cross Validation Logistic Regression Model Results: 3 Mins Before Eating}
\label{tbl:Logistic Regression_Eating_cv_results_min3}
\begin{tabular}{|p{1cm}| p{1cm} |p{1cm}| p{1cm}| p{1cm}| p{1cm}| p{1cm}| p{1cm} |p{1cm} |p{1cm} |p{1cm}| p{1cm} |p{1cm}|}
\toprule
\multirow{2}{*}{C} &\multicolumn{2}{c|}{fold1} & \multicolumn{2}{c|}{fold2} & \multicolumn{2}{c|}{fold3} & \multicolumn{2}{c|}{fold4} & \multicolumn{2}{c|}{fold5}  & \multicolumn{2}{c|}{average} \\\cline{2-13}
&\multicolumn{1}{c|}{Train} & \multicolumn{1}{c|}{Valid.} & \multicolumn{1}{c|}{Train} & \multicolumn{1}{c|}{Valid.} & \multicolumn{1}{c|}{Train}  & \multicolumn{1}{c|}{Valid.} &\multicolumn{1}{c|}{Train} & \multicolumn{1}{c|}{Valid.} & \multicolumn{1}{c|}{Train} & \multicolumn{1}{c|}{Valid.} & \multicolumn{1}{c|}{Train}  & \multicolumn{1}{c|}{Valid.}\\
\midrule
1000&0.6993&\textbf{0.6555}&0.6975&0.659&0.6824&0.6978&0.6888&0.6729&0.6955&0.6657&0.6927&0.6702\\
\midrule
100&0.7006&0.654&0.6975&0.6621&0.6822&0.6986&0.6887&0.6738&0.6956&0.6652&0.6929&0.6707\\
\midrule
10&0.705&0.6448&0.6934&0.6666&0.6813&0.6995&0.6852&0.6858&0.695&0.6665&0.692&0.6726\\
\midrule
1&0.7056&0.6397&0.6926&0.6645&0.6763&0.7014&0.6802&\textbf{0.6921}&0.6966&0.665&0.6903&0.6726\\
\midrule
0.1&0.7016&0.6464&0.6904&\textbf{0.6666}&0.6739&\textbf{0.7064}&0.6806&0.6919&0.6898&\textbf{0.6669}&0.6873&\textbf{0.6757}\\
\bottomrule
\end{tabular}
\end{table}

\begin{table}[ht!]
\centering
\caption{5-fold Cross Validation Logistic Regression Model Results: 4 Mins Before Eating}
\label{tbl:Logistic Regression_Eating_cv_results_min4}
\begin{tabular}{|p{1cm}| p{1cm} |p{1cm}| p{1cm}| p{1cm}| p{1cm}| p{1cm}| p{1cm} |p{1cm} |p{1cm} |p{1cm}| p{1cm} |p{1cm}|}
\toprule
\multirow{2}{*}{C} &\multicolumn{2}{c|}{fold1} & \multicolumn{2}{c|}{fold2} & \multicolumn{2}{c|}{fold3} & \multicolumn{2}{c|}{fold4} & \multicolumn{2}{c|}{fold5}  & \multicolumn{2}{c|}{average} \\\cline{2-13}
&\multicolumn{1}{c|}{Train} & \multicolumn{1}{c|}{Valid.} & \multicolumn{1}{c|}{Train} & \multicolumn{1}{c|}{Valid.} & \multicolumn{1}{c|}{Train}  & \multicolumn{1}{c|}{Valid.} &\multicolumn{1}{c|}{Train} & \multicolumn{1}{c|}{Valid.} & \multicolumn{1}{c|}{Train} & \multicolumn{1}{c|}{Valid.} & \multicolumn{1}{c|}{Train}  & \multicolumn{1}{c|}{Valid.}\\
\midrule
1000&0.6997&\textbf{0.6547}&0.6976&0.6607&0.682&0.6987&0.6872&0.6743&0.6966&0.6653&0.6926&0.6707\\
\midrule
100&0.6999&0.6543&0.6973&0.6608&0.6826&0.6995&0.6878&0.6768&0.6968&0.6657&0.6928&0.6714\\
\midrule
10&0.7045&0.6438&0.694&\textbf{0.6676}&0.6819&0.7009&0.6853&0.688&0.6962&0.6668&0.6924&0.6734\\
\midrule
1&0.7048&0.6394&0.6934&0.6667&0.6778&0.7019&0.681&\textbf{0.6933}&0.6977&\textbf{0.669}&0.6909&0.6741\\
\midrule
0.1&0.7026&0.646&0.6909&0.6666&0.6755&\textbf{0.7072}&0.6815&0.692&0.6895&0.6677&0.688&\textbf{0.6759}\\
\bottomrule
\end{tabular}
\end{table}

\begin{table}[ht!]
\centering
\caption{5-fold Cross Validation Logistic Regression Model Results: 0 Min Before Food Purchasing}
\label{tbl:Logistic Regression_Food Purchasing_cv_results_min0}
\begin{tabular}{|p{1cm}| p{1cm} |p{1cm}| p{1cm}| p{1cm}| p{1cm}| p{1cm}| p{1cm} |p{1cm} |p{1cm} |p{1cm}| p{1cm} |p{1cm}|}
\toprule
\multirow{2}{*}{C} &\multicolumn{2}{c|}{fold1} & \multicolumn{2}{c|}{fold2} & \multicolumn{2}{c|}{fold3} & \multicolumn{2}{c|}{fold4} & \multicolumn{2}{c|}{fold5}  & \multicolumn{2}{c|}{average} \\\cline{2-13}
&\multicolumn{1}{c|}{Train} & \multicolumn{1}{c|}{Valid.} & \multicolumn{1}{c|}{Train} & \multicolumn{1}{c|}{Valid.} & \multicolumn{1}{c|}{Train}  & \multicolumn{1}{c|}{Valid.} &\multicolumn{1}{c|}{Train} & \multicolumn{1}{c|}{Valid.} & \multicolumn{1}{c|}{Train} & \multicolumn{1}{c|}{Valid.} & \multicolumn{1}{c|}{Train}  & \multicolumn{1}{c|}{Valid.}\\
\midrule
1000&0.7603&0.5715&0.7751&0.7364&0.7988&0.5871&0.7746&0.6059&0.7453&0.7053&0.7708&0.6412\\
\midrule
100&0.7562&0.6304&0.7804&0.7273&0.7863&0.6132&0.7705&0.5894&0.7313&0.712&0.7649&0.6545\\
\midrule
10&0.7314&0.6373&0.7566&0.725&0.7759&0.6217&0.7705&0.6028&0.7243&0.6903&0.7517&0.6554\\
\midrule
1&0.7273&\textbf{0.6634}&0.7513&0.7309&0.7676&\textbf{0.6268}&0.75&0.6045&0.7126&0.7099&0.7418&0.6671\\
\midrule
0.1&0.7066&0.6414&0.7434&\textbf{0.7432}&0.7469&0.6222&0.7295&\textbf{0.6424}&0.722&\textbf{0.7156}&0.7297&\textbf{0.6729}\\
\bottomrule
\end{tabular}
\end{table}

\begin{table}[ht!]
\centering
\caption{5-fold Cross Validation Logistic Regression Model Results: 1 Min Before Food Purchasing}
\label{tbl:Logistic Regression_Food Purchasing_cv_results_min1}
\begin{tabular}{|p{1cm}| p{1cm} |p{1cm}| p{1cm}| p{1cm}| p{1cm}| p{1cm}| p{1cm} |p{1cm} |p{1cm} |p{1cm}| p{1cm} |p{1cm}|}
\toprule
\multirow{2}{*}{C} &\multicolumn{2}{c|}{fold1} & \multicolumn{2}{c|}{fold2} & \multicolumn{2}{c|}{fold3} & \multicolumn{2}{c|}{fold4} & \multicolumn{2}{c|}{fold5}  & \multicolumn{2}{c|}{average} \\\cline{2-13}
&\multicolumn{1}{c|}{Train} & \multicolumn{1}{c|}{Valid.} & \multicolumn{1}{c|}{Train} & \multicolumn{1}{c|}{Valid.} & \multicolumn{1}{c|}{Train}  & \multicolumn{1}{c|}{Valid.} &\multicolumn{1}{c|}{Train} & \multicolumn{1}{c|}{Valid.} & \multicolumn{1}{c|}{Train} & \multicolumn{1}{c|}{Valid.} & \multicolumn{1}{c|}{Train}  & \multicolumn{1}{c|}{Valid.}\\
\midrule
1000&0.7727&0.5813&0.7831&0.7245&0.7925&0.5954&0.7725&0.5975&0.7336&\textbf{0.7321}&0.7709&0.6461\\
\midrule
100&0.7541&0.6042&0.7751&0.7314&0.7801&0.6225&0.7705&0.5987&0.729&0.7139&0.7618&0.6541\\
\midrule
10&0.7355&0.6294&0.7646&0.7233&0.7822&0.6216&0.7705&0.6114&0.7266&0.6898&0.7559&0.6551\\
\midrule
1&0.7107&\textbf{0.6954}&0.7619&0.7324&0.7573&0.61&0.7561&0.603&0.729&0.7133&0.743&0.6708\\
\midrule
0.1&0.7128&0.6695&0.7381&\textbf{0.7424}&0.7573&\textbf{0.6374}&0.7295&\textbf{0.6593}&0.7266&0.707&0.7329&\textbf{0.6831}\\
\bottomrule
\end{tabular}
\end{table}

\begin{table}[ht!]
\centering
\caption{5-fold Cross Validation Logistic Regression Model Results: 2 Mins Before Food Purchasing}
\label{tbl:Logistic Regression_Food Purchasing_cv_results_min2}
\begin{tabular}{|p{1cm}| p{1cm} |p{1cm}| p{1cm}| p{1cm}| p{1cm}| p{1cm}| p{1cm} |p{1cm} |p{1cm} |p{1cm}| p{1cm} |p{1cm}|}
\toprule
\multirow{2}{*}{C} &\multicolumn{2}{c|}{fold1} & \multicolumn{2}{c|}{fold2} & \multicolumn{2}{c|}{fold3} & \multicolumn{2}{c|}{fold4} & \multicolumn{2}{c|}{fold5}  & \multicolumn{2}{c|}{average} \\\cline{2-13}
&\multicolumn{1}{c|}{Train} & \multicolumn{1}{c|}{Valid.} & \multicolumn{1}{c|}{Train} & \multicolumn{1}{c|}{Valid.} & \multicolumn{1}{c|}{Train}  & \multicolumn{1}{c|}{Valid.} &\multicolumn{1}{c|}{Train} & \multicolumn{1}{c|}{Valid.} & \multicolumn{1}{c|}{Train} & \multicolumn{1}{c|}{Valid.} & \multicolumn{1}{c|}{Train}  & \multicolumn{1}{c|}{Valid.}\\
\midrule
1000&0.7831&0.5999&0.7751&0.7359&0.7967&0.5874&0.7684&0.6209&0.75&\textbf{0.7269}&0.7747&0.6542\\
\midrule
100&0.7665&0.6162&0.7778&\textbf{0.7444}&0.7967&0.6065&0.7684&0.6058&0.75&0.7196&0.7719&0.6585\\
\midrule
10&0.7417&0.6471&0.7619&0.7363&0.7822&0.5966&0.7684&0.6104&0.7407&0.7011&0.759&0.6583\\
\midrule
1&0.7128&\textbf{0.6873}&0.7619&0.7313&0.7676&0.6027&0.7561&0.612&0.7336&0.7129&0.7464&0.6692\\
\midrule
0.1&0.7149&0.6757&0.7354&0.7427&0.7614&\textbf{0.6132}&0.7357&\textbf{0.6507}&0.7266&0.7184&0.7348&\textbf{0.6801}\\
\bottomrule
\end{tabular}
\end{table}

\begin{table}[ht!]
\centering
\caption{5-fold Cross Validation Logistic Regression Model Results: 3 Mins Before Food Purchasing}
\label{tbl:Logistic Regression_Food Purchasing_cv_results_min3}
\begin{tabular}{|p{1cm}| p{1cm} |p{1cm}| p{1cm}| p{1cm}| p{1cm}| p{1cm}| p{1cm} |p{1cm} |p{1cm} |p{1cm}| p{1cm} |p{1cm}|}
\toprule
\multirow{2}{*}{C} &\multicolumn{2}{c|}{fold1} & \multicolumn{2}{c|}{fold2} & \multicolumn{2}{c|}{fold3} & \multicolumn{2}{c|}{fold4} & \multicolumn{2}{c|}{fold5}  & \multicolumn{2}{c|}{average} \\\cline{2-13}
&\multicolumn{1}{c|}{Train} & \multicolumn{1}{c|}{Valid.} & \multicolumn{1}{c|}{Train} & \multicolumn{1}{c|}{Valid.} & \multicolumn{1}{c|}{Train}  & \multicolumn{1}{c|}{Valid.} &\multicolumn{1}{c|}{Train} & \multicolumn{1}{c|}{Valid.} & \multicolumn{1}{c|}{Train} & \multicolumn{1}{c|}{Valid.} & \multicolumn{1}{c|}{Train}  & \multicolumn{1}{c|}{Valid.}\\
\midrule
1000&0.7831&0.6175&0.7751&0.7205&0.8008&0.5874&0.7807&0.6226&0.7453&0.7087&0.777&0.6513\\
\midrule
100&0.7665&0.6146&0.7778&0.7269&0.7946&0.5877&0.7787&0.6147&0.7383&0.6944&0.7712&0.6477\\
\midrule
10&0.7397&0.6535&0.7672&0.7168&0.7759&0.6035&0.7766&0.6105&0.7266&0.6833&0.7572&0.6535\\
\midrule
1&0.7211&\textbf{0.6791}&0.7513&0.7268&0.7614&0.6023&0.7623&0.6208&0.7336&0.7071&0.7459&0.6672\\
\midrule
0.1&0.7169&0.6753&0.7328&\textbf{0.7429}&0.7614&\textbf{0.6132}&0.7377&\textbf{0.6416}&0.722&\textbf{0.7184}&0.7342&\textbf{0.6783}\\
\bottomrule
\end{tabular}
\end{table}

\begin{table}[ht!]
\centering
\caption{5-fold Cross Validation Logistic Regression Model Results: 4 Mins Before Food Purchasing}
\label{tbl:Logistic Regression_Food Purchasing_cv_results_min4}
\begin{tabular}{|p{1cm}| p{1cm} |p{1cm}| p{1cm}| p{1cm}| p{1cm}| p{1cm}| p{1cm} |p{1cm} |p{1cm} |p{1cm}| p{1cm} |p{1cm}|}
\toprule
\multirow{2}{*}{C} &\multicolumn{2}{c|}{fold1} & \multicolumn{2}{c|}{fold2} & \multicolumn{2}{c|}{fold3} & \multicolumn{2}{c|}{fold4} & \multicolumn{2}{c|}{fold5}  & \multicolumn{2}{c|}{average} \\\cline{2-13}
&\multicolumn{1}{c|}{Train} & \multicolumn{1}{c|}{Valid.} & \multicolumn{1}{c|}{Train} & \multicolumn{1}{c|}{Valid.} & \multicolumn{1}{c|}{Train}  & \multicolumn{1}{c|}{Valid.} &\multicolumn{1}{c|}{Train} & \multicolumn{1}{c|}{Valid.} & \multicolumn{1}{c|}{Train} & \multicolumn{1}{c|}{Valid.} & \multicolumn{1}{c|}{Train}  & \multicolumn{1}{c|}{Valid.}\\
\midrule
1000&0.7665&0.5788&0.7778&0.7146&0.7925&0.6039&0.7766&0.5613&0.7383&0.7016&0.7704&0.632\\
\midrule
100&0.7541&0.6294&0.7725&0.7332&0.7925&0.5711&0.7766&0.589&0.7336&\textbf{0.7268}&0.7659&0.6499\\
\midrule
10&0.7397&0.6432&0.7646&0.7259&0.7697&\textbf{0.6116}&0.7705&0.6017&0.7313&0.6937&0.7551&0.6552\\
\midrule
1&0.719&\textbf{0.6866}&0.7513&0.7228&0.7676&0.6107&0.7623&0.612&0.7313&0.7122&0.7463&0.6689\\
\midrule
0.1&0.7128&0.675&0.7328&\textbf{0.739}&0.7656&0.6052&0.7336&\textbf{0.6593}&0.7243&0.7179&0.7338&\textbf{0.6793}\\
\bottomrule
\end{tabular}
\end{table}

\begin{table}[ht!]
\centering
\caption{5-fold Cross Validation RBF-SVM Model Results: 0 Min Before Eating}
\label{tbl:RBF-SVM_Eating_cv_results_min0}
\begin{tabular}{|c|c|c|c|c|c|c|c|c|c|c|c|c|c|}
\toprule
\multirow{2}{*}{C } & \multirow{2}{*}{$\gamma$} &\multicolumn{2}{c|}{fold1} & \multicolumn{2}{c|}{fold2} & \multicolumn{2}{c|}{fold3} & \multicolumn{2}{c|}{fold4} & \multicolumn{2}{c|}{fold5}  & \multicolumn{2}{c|}{average} \\\cline{3-14}
&&\multicolumn{1}{c|}{Train} & \multicolumn{1}{c|}{Valid.} & \multicolumn{1}{c|}{Train} & \multicolumn{1}{c|}{Valid.} & \multicolumn{1}{c|}{Train}  & \multicolumn{1}{c|}{Valid.} &\multicolumn{1}{c|}{Train} & \multicolumn{1}{c|}{Valid.} & \multicolumn{1}{c|}{Train} & \multicolumn{1}{c|}{Valid.} & \multicolumn{1}{c|}{Train}  & \multicolumn{1}{c|}{Valid.}\\        \midrule
1000&0.01&0.5811&0.5404&0.5261&0.5618&0.5909&0.5607&0.6408&0.5992&0.5148&0.5223&0.5707&0.5569\\[6pt]
\midrule
1000&0.1&0.5109&0.5067&0.5745&0.5457&0.5732&0.576&0.5914&0.5943&0.4821&0.4677&0.5464&0.5381\\[6pt]
\midrule
1000&1&0.548&0.5289&0.6218&0.5565&0.5568&0.5215&0.5824&0.526&0.5478&0.5304&0.5714&0.5326\\[6pt]
\midrule
1000&10&0.5497&0.5008&0.5962&0.5198&0.557&0.4914&0.5527&0.5134&0.5613&0.5212&0.5634&0.5093\\[6pt]
\midrule
1000&100&0.7106&0.5156&0.6724&0.505&0.6575&0.5209&0.713&0.5256&0.7148&0.52&0.6937&0.5174\\[6pt]
\midrule
100&0.01&0.5184&0.5064&0.6585&\textbf{0.6573}&0.5628&0.5685&0.6196&\textbf{0.6433}&0.6167&0.6317&0.5952&0.6014\\
\bottomrule
\end{tabular}
\end{table}

\begin{table}[ht!]
\centering
\caption{5-fold Cross Validation RBF-SVM Model Results: 1 Min Before Eating}
\label{tbl:RBF-SVM_Eating_cv_results_min1}
\begin{tabular}{|c|c|c|c|c|c|c|c|c|c|c|c|c|c|}
\toprule
\multirow{2}{*}{C } & \multirow{2}{*}{$\gamma$} &\multicolumn{2}{c|}{fold1} & \multicolumn{2}{c|}{fold2} & \multicolumn{2}{c|}{fold3} & \multicolumn{2}{c|}{fold4} & \multicolumn{2}{c|}{fold5}  & \multicolumn{2}{c|}{average} \\\cline{3-14}
&&\multicolumn{1}{c|}{Train} & \multicolumn{1}{c|}{Valid.} & \multicolumn{1}{c|}{Train} & \multicolumn{1}{c|}{Valid.} & \multicolumn{1}{c|}{Train}  & \multicolumn{1}{c|}{Valid.} &\multicolumn{1}{c|}{Train} & \multicolumn{1}{c|}{Valid.} & \multicolumn{1}{c|}{Train} & \multicolumn{1}{c|}{Valid.} & \multicolumn{1}{c|}{Train}  & \multicolumn{1}{c|}{Valid.}\\        \midrule
1000&0.01&0.5814&0.5338&0.5665&0.5787&0.5234&0.5417&0.4886&0.5151&0.4756&0.4641&0.5271&0.5267\\[6pt]
\midrule
1000&0.1&0.5604&0.5249&0.6024&0.5783&0.6174&0.6311&0.5093&0.5334&0.5493&0.5508&0.5678&0.5637\\[6pt]
\midrule
1000&1&0.583&0.5358&0.5308&0.5162&0.5737&0.4963&0.6043&0.5343&0.5697&0.558&0.5723&0.5281\\[6pt]
\midrule
1000&10&0.5622&0.5039&0.6944&0.5395&0.5475&0.4935&0.5443&0.5038&0.5593&0.5136&0.5816&0.5108\\[6pt]
\midrule
1000&100&0.7&0.5115&0.7247&0.5119&0.6946&0.5365&0.7095&0.5233&0.7301&0.516&0.7118&0.5198\\[6pt]
\midrule
100&0.01&0.614&0.5774&0.5644&0.5556&0.563&0.5929&0.5241&0.5543&0.5907&0.6023&0.5712&0.5765\\
\bottomrule
\end{tabular}
\end{table}

\begin{table}[ht!]
\centering
\caption{5-fold Cross Validation RBF-SVM Model Results: 2 Mins Before Eating}
\label{tbl:RBF-SVM_Eating_cv_results_min2}
\begin{tabular}{|c|c|c|c|c|c|c|c|c|c|c|c|c|c|}
\toprule
\multirow{2}{*}{C } & \multirow{2}{*}{$\gamma$} &\multicolumn{2}{c|}{fold1} & \multicolumn{2}{c|}{fold2} & \multicolumn{2}{c|}{fold3} & \multicolumn{2}{c|}{fold4} & \multicolumn{2}{c|}{fold5}  & \multicolumn{2}{c|}{average} \\\cline{3-14}
&&\multicolumn{1}{c|}{Train} & \multicolumn{1}{c|}{Valid.} & \multicolumn{1}{c|}{Train} & \multicolumn{1}{c|}{Valid.} & \multicolumn{1}{c|}{Train}  & \multicolumn{1}{c|}{Valid.} &\multicolumn{1}{c|}{Train} & \multicolumn{1}{c|}{Valid.} & \multicolumn{1}{c|}{Train} & \multicolumn{1}{c|}{Valid.} & \multicolumn{1}{c|}{Train}  & \multicolumn{1}{c|}{Valid.}\\        \midrule
1000&0.01&0.6087&0.5674&0.5908&0.6099&0.5317&0.5348&0.5478&0.5435&0.5244&0.537&0.5607&0.5585\\[6pt]
\midrule
1000&0.1&0.6272&0.5608&0.624&0.6063&0.5295&0.5831&0.5835&0.6208&0.5711&0.5951&0.5871&0.5932\\[6pt]
\midrule
1000&1&0.5901&0.5139&0.5874&0.55&0.5665&0.4885&0.5635&0.524&0.5708&0.5557&0.5757&0.5264\\[6pt]
\midrule
1000&10&0.553&0.5008&0.6429&0.5255&0.5563&0.4964&0.5575&0.4869&0.5586&0.5245&0.5736&0.5068\\[6pt]
\midrule
1000&100&0.6977&0.5095&0.7058&0.5093&0.6868&0.516&0.7203&0.5268&0.7219&0.5212&0.7065&0.5165\\[6pt]
\midrule
100&0.01&0.6506&0.6113&0.5907&\textbf{0.6139}&0.6091&0.6462&0.5893&0.5599&0.6664&\textbf{0.6315}&0.6212&\textbf{0.6126}\\
\bottomrule
\end{tabular}
\end{table}

\begin{table}[ht!]
\centering
\caption{5-fold Cross Validation RBF-SVM Model Results: 3 Mins Before Eating}
\label{tbl:RBF-SVM_Eating_cv_results_min3}
\begin{tabular}{|c|c|c|c|c|c|c|c|c|c|c|c|c|c|}
\toprule
\multirow{2}{*}{C } & \multirow{2}{*}{$\gamma$} &\multicolumn{2}{c|}{fold1} & \multicolumn{2}{c|}{fold2} & \multicolumn{2}{c|}{fold3} & \multicolumn{2}{c|}{fold4} & \multicolumn{2}{c|}{fold5}  & \multicolumn{2}{c|}{average} \\\cline{3-14}
&&\multicolumn{1}{c|}{Train} & \multicolumn{1}{c|}{Valid.} & \multicolumn{1}{c|}{Train} & \multicolumn{1}{c|}{Valid.} & \multicolumn{1}{c|}{Train}  & \multicolumn{1}{c|}{Valid.} &\multicolumn{1}{c|}{Train} & \multicolumn{1}{c|}{Valid.} & \multicolumn{1}{c|}{Train} & \multicolumn{1}{c|}{Valid.} & \multicolumn{1}{c|}{Train}  & \multicolumn{1}{c|}{Valid.}\\        \midrule
1000&0.01&0.5831&0.5409&0.6327&0.6541&0.556&0.5697&0.5255&0.5511&0.4722&0.4831&0.5539&0.5598\\[6pt]
\midrule
1000&0.1&0.6112&0.597&0.5598&0.5677&0.5586&0.5641&0.6153&0.5746&0.5389&0.5164&0.5768&0.564\\[6pt]
\midrule
1000&1&0.5544&0.5368&0.5805&0.5184&0.5714&0.5108&0.5517&0.5101&0.5683&0.5426&0.5653&0.5237\\[6pt]
\midrule
1000&10&0.5554&0.5091&0.684&0.5331&0.5435&0.4928&0.5492&0.521&0.5499&0.5197&0.5764&0.5151\\[6pt]
\midrule
1000&100&0.7095&0.5158&0.7259&0.5114&0.6791&0.5187&0.7375&0.5239&0.7289&0.522&0.7162&0.5184\\[6pt]
\midrule
100&0.01&0.5874&0.5614&0.5126&0.5224&0.5154&0.5288&0.6199&0.5925&0.5346&0.5366&0.554&0.5483\\
\bottomrule
\end{tabular}
\end{table}

\begin{table}[ht!]
\centering
\caption{5-fold Cross Validation RBF-SVM Model Results: 4 Mins Before Eating}
\label{tbl:RBF-SVM_Eating_cv_results_min4}
\begin{tabular}{|c|c|c|c|c|c|c|c|c|c|c|c|c|c|}
\toprule
\multirow{2}{*}{C } & \multirow{2}{*}{$\gamma$} &\multicolumn{2}{c|}{fold1} & \multicolumn{2}{c|}{fold2} & \multicolumn{2}{c|}{fold3} & \multicolumn{2}{c|}{fold4} & \multicolumn{2}{c|}{fold5}  & \multicolumn{2}{c|}{average} \\\cline{3-14}
&&\multicolumn{1}{c|}{Train} & \multicolumn{1}{c|}{Valid.} & \multicolumn{1}{c|}{Train} & \multicolumn{1}{c|}{Valid.} & \multicolumn{1}{c|}{Train}  & \multicolumn{1}{c|}{Valid.} &\multicolumn{1}{c|}{Train} & \multicolumn{1}{c|}{Valid.} & \multicolumn{1}{c|}{Train} & \multicolumn{1}{c|}{Valid.} & \multicolumn{1}{c|}{Train}  & \multicolumn{1}{c|}{Valid.}\\        \midrule
1000&0.01&0.6331&0.5883&0.6107&0.6001&0.614&\textbf{0.6347}&0.5012&0.4906&0.5295&0.5153&0.5777&0.5658\\[6pt]
\midrule
1000&0.1&0.6189&0.58&0.6281&0.595&0.5934&0.5646&0.6197&0.6013&0.6324&0.6028&0.6185&0.5888\\[6pt]
\midrule
1000&1&0.5924&0.5181&0.5448&0.5253&0.5552&0.5126&0.5635&0.5395&0.5601&0.5503&0.5632&0.5292\\[6pt]
\midrule
1000&10&0.5501&0.5063&0.6684&0.5409&0.5385&0.4914&0.6492&0.5266&0.5379&0.5115&0.5888&0.5153\\[6pt]
\midrule
1000&100&0.6971&0.5105&0.6897&0.5098&0.6697&0.5272&0.7307&0.5293&0.7118&0.5196&0.6998&0.5193\\[6pt]
\midrule
100&0.01&0.6367&\textbf{0.6195}&0.5733&0.573&0.5933&0.5724&0.6137&\textbf{0.6463}&0.617&0.633&0.6068&\textbf{0.6088}\\
\bottomrule
\end{tabular}
\end{table}

\begin{table}[ht!]
\centering
\caption{5-fold Cross Validation RBF-SVM Model Results: 0 Min Before Food Purchasing}
\label{tbl:RBF-SVM_Food Purchasing_cv_results_min0}
\begin{tabular}{|c|c|c|c|c|c|c|c|c|c|c|c|c|c|}
\toprule
\multirow{2}{*}{C } & \multirow{2}{*}{$\gamma$} &\multicolumn{2}{c|}{fold1} & \multicolumn{2}{c|}{fold2} & \multicolumn{2}{c|}{fold3} & \multicolumn{2}{c|}{fold4} & \multicolumn{2}{c|}{fold5}  & \multicolumn{2}{c|}{average} \\\cline{3-14}
&&\multicolumn{1}{c|}{Train} & \multicolumn{1}{c|}{Valid.} & \multicolumn{1}{c|}{Train} & \multicolumn{1}{c|}{Valid.} & \multicolumn{1}{c|}{Train}  & \multicolumn{1}{c|}{Valid.} &\multicolumn{1}{c|}{Train} & \multicolumn{1}{c|}{Valid.} & \multicolumn{1}{c|}{Train} & \multicolumn{1}{c|}{Valid.} & \multicolumn{1}{c|}{Train}  & \multicolumn{1}{c|}{Valid.}\\        \midrule
1000&0.01&0.8244&0.6544&0.8571&0.6931&0.8257&0.639&0.8135&0.6008&0.8294&0.6185&0.83&0.6412\\[6pt]
\midrule
1000&0.1&0.9855&0.5362&0.9894&0.5941&0.9627&0.5743&0.959&0.6354&0.965&0.6406&0.9723&0.5961\\[6pt]
\midrule
1000&1&1.0&0.5499&1.0&0.5488&0.9959&0.6205&0.9959&0.6702&1.0&0.5988&0.9984&0.5976\\[6pt]
\midrule
1000&10&1.0&0.5611&1.0&0.579&1.0&0.5527&1.0&0.6063&1.0&0.5525&1.0&0.5703\\[6pt]
\midrule
1000&100&1.0&0.493&1.0&0.497&1.0&0.4885&1.0&0.5177&1.0&0.4976&1.0&0.4988\\[6pt]
\midrule
100&0.01&0.7479&0.698&0.7698&0.7272&0.7697&0.642&0.7725&0.6217&0.757&0.7275&0.7634&0.6833\\
\bottomrule
\end{tabular}
\end{table}

\begin{table}[ht!]
\centering
\caption{5-fold Cross Validation RBF-SVM Model Results: 1 Min Before Food Purchasing}
\label{tbl:RBF-SVM_Food Purchasing_cv_results_min1}
\begin{tabular}{|c|c|c|c|c|c|c|c|c|c|c|c|c|c|}
\toprule
\multirow{2}{*}{C } & \multirow{2}{*}{$\gamma$} &\multicolumn{2}{c|}{fold1} & \multicolumn{2}{c|}{fold2} & \multicolumn{2}{c|}{fold3} & \multicolumn{2}{c|}{fold4} & \multicolumn{2}{c|}{fold5}  & \multicolumn{2}{c|}{average} \\\cline{3-14}
&&\multicolumn{1}{c|}{Train} & \multicolumn{1}{c|}{Valid.} & \multicolumn{1}{c|}{Train} & \multicolumn{1}{c|}{Valid.} & \multicolumn{1}{c|}{Train}  & \multicolumn{1}{c|}{Valid.} &\multicolumn{1}{c|}{Train} & \multicolumn{1}{c|}{Valid.} & \multicolumn{1}{c|}{Train} & \multicolumn{1}{c|}{Valid.} & \multicolumn{1}{c|}{Train}  & \multicolumn{1}{c|}{Valid.}\\        \midrule
1000&0.01&0.814&0.6554&0.8386&0.7076&0.832&0.6199&0.8033&0.6048&0.8271&0.6138&0.823&0.6403\\[6pt]
\midrule
1000&0.1&0.9876&0.5406&0.9788&0.5733&0.9751&0.5734&0.9652&0.6139&0.9509&0.5812&0.9715&0.5765\\[6pt]
\midrule
1000&1&1.0&0.5471&1.0&0.5595&0.9959&0.5912&0.9959&0.6207&1.0&0.6192&0.9984&0.5875\\[6pt]
\midrule
1000&10&1.0&0.5631&1.0&0.5864&1.0&0.5576&1.0&0.6128&1.0&0.5633&1.0&0.5766\\[6pt]
\midrule
1000&100&1.0&0.5017&1.0&0.5054&1.0&0.4966&1.0&0.5171&1.0&0.5147&1.0&0.5071\\[6pt]
\midrule
100&0.01&0.7417&0.6903&0.7725&0.7275&0.7718&0.6453&0.7746&0.6123&0.7547&0.7291&0.7631&0.6809\\
\bottomrule
\end{tabular}
\end{table}

\begin{table}[ht!]
\centering
\caption{5-fold Cross Validation RBF-SVM Model Results: 2 Mins Before Food Purchasing}
\label{tbl:RBF-SVM_Food Purchasing_cv_results_min2}
\begin{tabular}{|c|c|c|c|c|c|c|c|c|c|c|c|c|c|}
\toprule
\multirow{2}{*}{C } & \multirow{2}{*}{$\gamma$} &\multicolumn{2}{c|}{fold1} & \multicolumn{2}{c|}{fold2} & \multicolumn{2}{c|}{fold3} & \multicolumn{2}{c|}{fold4} & \multicolumn{2}{c|}{fold5}  & \multicolumn{2}{c|}{average} \\\cline{3-14}
&&\multicolumn{1}{c|}{Train} & \multicolumn{1}{c|}{Valid.} & \multicolumn{1}{c|}{Train} & \multicolumn{1}{c|}{Valid.} & \multicolumn{1}{c|}{Train}  & \multicolumn{1}{c|}{Valid.} &\multicolumn{1}{c|}{Train} & \multicolumn{1}{c|}{Valid.} & \multicolumn{1}{c|}{Train} & \multicolumn{1}{c|}{Valid.} & \multicolumn{1}{c|}{Train}  & \multicolumn{1}{c|}{Valid.}\\        \midrule
1000&0.01&0.8182&0.6578&0.8492&0.7184&0.8402&0.6299&0.8074&0.632&0.8364&0.6335&0.8303&0.6543\\[6pt]
\midrule
1000&0.1&0.9917&0.5497&0.9762&0.5782&0.9689&0.5836&0.957&0.6452&0.9579&0.575&0.9703&0.5863\\[6pt]
\midrule
1000&1&1.0&0.543&0.9974&0.5702&0.9979&0.6347&0.9939&0.5998&0.9977&0.631&0.9974&0.5957\\[6pt]
\midrule
1000&10&1.0&0.5575&1.0&0.5928&1.0&0.5994&1.0&0.6115&1.0&0.5498&1.0&0.5822\\[6pt]
\midrule
1000&100&1.0&0.5018&1.0&0.4966&1.0&0.4888&1.0&0.5091&1.0&0.4998&1.0&0.4992\\[6pt]
\midrule
100&0.01&0.7521&0.6976&0.7698&0.7273&0.7739&0.6503&0.7766&0.6115&0.7523&0.7431&0.7649&0.686\\
\bottomrule
\end{tabular}
\end{table}

\begin{table}[ht!]
\centering
\caption{5-fold Cross Validation RBF-SVM Model Results: 3 Mins Before Food Purchasing}
\label{tbl:RBF-SVM_Food Purchasing_cv_results_min3}
\begin{tabular}{|c|c|c|c|c|c|c|c|c|c|c|c|c|c|}
\toprule
\multirow{2}{*}{C } & \multirow{2}{*}{$\gamma$} &\multicolumn{2}{c|}{fold1} & \multicolumn{2}{c|}{fold2} & \multicolumn{2}{c|}{fold3} & \multicolumn{2}{c|}{fold4} & \multicolumn{2}{c|}{fold5}  & \multicolumn{2}{c|}{average} \\\cline{3-14}
&&\multicolumn{1}{c|}{Train} & \multicolumn{1}{c|}{Valid.} & \multicolumn{1}{c|}{Train} & \multicolumn{1}{c|}{Valid.} & \multicolumn{1}{c|}{Train}  & \multicolumn{1}{c|}{Valid.} &\multicolumn{1}{c|}{Train} & \multicolumn{1}{c|}{Valid.} & \multicolumn{1}{c|}{Train} & \multicolumn{1}{c|}{Valid.} & \multicolumn{1}{c|}{Train}  & \multicolumn{1}{c|}{Valid.}\\        \midrule
1000&0.01&0.8223&0.6576&0.8466&0.7155&0.834&0.6273&0.8074&0.6134&0.8411&0.6041&0.8303&0.6436\\[6pt]
\midrule
1000&0.1&0.9855&0.5575&0.9841&0.593&0.9689&0.6051&0.9631&0.6404&0.9533&0.5796&0.971&0.5951\\[6pt]
\midrule
1000&1&0.9979&0.5587&0.9974&0.5849&0.9979&0.6475&0.998&0.6139&0.9977&0.6371&0.9978&0.6084\\[6pt]
\midrule
1000&10&1.0&0.5195&1.0&0.5983&1.0&0.5509&1.0&0.5817&1.0&0.5342&1.0&0.5569\\[6pt]
\midrule
1000&100&1.0&0.4942&1.0&0.4999&1.0&0.4901&1.0&0.5085&1.0&0.5063&1.0&0.4998\\[6pt]
\midrule
100&0.01&0.7438&0.6993&0.7593&0.7313&0.7697&0.651&0.7746&0.6124&0.75&0.7383&0.7595&0.6865\\
\bottomrule
\end{tabular}
\end{table}

\begin{table}[ht!]
\centering
\caption{5-fold Cross Validation RBF-SVM Model Results: 4 Mins Before Food Purchasing}
\label{tbl:RBF-SVM_Food Purchasing_cv_results_min4}
\begin{tabular}{|c|c|c|c|c|c|c|c|c|c|c|c|c|c|}
\toprule
\multirow{2}{*}{C } & \multirow{2}{*}{$\gamma$} &\multicolumn{2}{c|}{fold1} & \multicolumn{2}{c|}{fold2} & \multicolumn{2}{c|}{fold3} & \multicolumn{2}{c|}{fold4} & \multicolumn{2}{c|}{fold5}  & \multicolumn{2}{c|}{average} \\\cline{3-14}
&&\multicolumn{1}{c|}{Train} & \multicolumn{1}{c|}{Valid.} & \multicolumn{1}{c|}{Train} & \multicolumn{1}{c|}{Valid.} & \multicolumn{1}{c|}{Train}  & \multicolumn{1}{c|}{Valid.} &\multicolumn{1}{c|}{Train} & \multicolumn{1}{c|}{Valid.} & \multicolumn{1}{c|}{Train} & \multicolumn{1}{c|}{Valid.} & \multicolumn{1}{c|}{Train}  & \multicolumn{1}{c|}{Valid.}\\        \midrule
1000&0.01&0.8326&0.6636&0.8439&0.7104&0.8216&0.6205&0.8115&0.5828&0.8154&0.6022&0.825&0.6359\\[6pt]
\midrule
1000&0.1&0.9897&0.589&0.9788&0.5844&0.9668&0.5735&0.9529&0.6401&0.9533&0.5628&0.9683&0.5899\\[6pt]
\midrule
1000&1&1.0&0.5632&0.9974&0.5625&0.9979&0.6704&0.9959&0.6439&1.0&0.5547&0.9982&0.599\\[6pt]
\midrule
1000&10&1.0&0.5294&1.0&0.5717&1.0&0.5883&1.0&0.6316&1.0&0.5241&1.0&0.569\\[6pt]
\midrule
1000&100&1.0&0.4939&1.0&0.5045&1.0&0.5213&1.0&0.517&1.0&0.5043&1.0&0.5082\\[6pt]
\midrule
100&0.01&0.7376&0.7013&0.7619&0.7296&0.7739&0.6515&0.7725&0.6201&0.7547&0.7254&0.7601&0.6856\\
\bottomrule
\end{tabular}
\end{table}

\begin{table}[ht!]
\centering
\caption{5-fold Cross Validation Random Forest Model Results: 0 Min Before Eating}
\label{tbl:Random Forest_Eating_cv_results_min0}
\begin{tabular}{|c|c|c|c|c|c|c|c|c|c|c|c|c|c|}
\toprule
\multirow{2}{*}{Tree Depth} & \multirow{2}{*}{Num. Tree} &\multicolumn{2}{c|}{fold1} & \multicolumn{2}{c|}{fold2} & \multicolumn{2}{c|}{fold3} & \multicolumn{2}{c|}{fold4} & \multicolumn{2}{c|}{fold5}  & \multicolumn{2}{c|}{average} \\\cline{3-14}
&&\multicolumn{1}{c|}{Train} & \multicolumn{1}{c|}{Valid.} & \multicolumn{1}{c|}{Train} & \multicolumn{1}{c|}{Valid.} & \multicolumn{1}{c|}{Train}  & \multicolumn{1}{c|}{Valid.} &\multicolumn{1}{c|}{Train} & \multicolumn{1}{c|}{Valid.} & \multicolumn{1}{c|}{Train} & \multicolumn{1}{c|}{Valid.} & \multicolumn{1}{c|}{Train}  & \multicolumn{1}{c|}{Valid.}\\        \midrule
2&10&0.6734&0.6421&0.6886&0.682&0.6694&0.6603&0.6828&0.6834&0.686&0.6548&0.6801&0.6645\\[6pt]
\midrule
2&30&0.709&0.6593&0.6959&0.674&0.6921&0.6906&0.6824&0.6868&0.7051&0.6489&0.6969&0.6719\\[6pt]
\midrule
2&50&0.7136&0.6499&0.6994&0.6812&0.6997&0.7073&0.6904&0.6851&0.7062&0.6503&0.7019&0.6748\\[6pt]
\midrule
2&100&0.7101&0.653&0.692&\textbf{0.6971}&0.6995&0.6904&0.6882&0.6948&0.7085&0.6623&0.6997&0.6795\\[6pt]
\midrule
2&200&0.7088&0.6493&0.695&0.6885&0.6893&0.6751&0.6865&0.6961&0.7035&0.66&0.6966&0.6738\\[6pt]
\midrule
5&10&0.7445&0.6675&0.7412&0.6809&0.7202&0.7195&0.7333&\textbf{0.7161}&0.7436&0.6645&0.7366&\textbf{0.6897}\\
\bottomrule
\end{tabular}
\end{table}

\begin{table}[ht!]
\centering
\caption{5-fold Cross Validation Random Forest Model Results: 1 Min Before Eating}
\label{tbl:Random Forest_Eating_cv_results_min1}
\begin{tabular}{|c|c|c|c|c|c|c|c|c|c|c|c|c|c|}
\toprule
\multirow{2}{*}{Tree Depth} & \multirow{2}{*}{Num. Tree} &\multicolumn{2}{c|}{fold1} & \multicolumn{2}{c|}{fold2} & \multicolumn{2}{c|}{fold3} & \multicolumn{2}{c|}{fold4} & \multicolumn{2}{c|}{fold5}  & \multicolumn{2}{c|}{average} \\\cline{3-14}
&&\multicolumn{1}{c|}{Train} & \multicolumn{1}{c|}{Valid.} & \multicolumn{1}{c|}{Train} & \multicolumn{1}{c|}{Valid.} & \multicolumn{1}{c|}{Train}  & \multicolumn{1}{c|}{Valid.} &\multicolumn{1}{c|}{Train} & \multicolumn{1}{c|}{Valid.} & \multicolumn{1}{c|}{Train} & \multicolumn{1}{c|}{Valid.} & \multicolumn{1}{c|}{Train}  & \multicolumn{1}{c|}{Valid.}\\        \midrule
2&10&0.6725&0.6425&0.6895&0.6802&0.6692&0.6414&0.6544&0.6722&0.6859&0.6554&0.6743&0.6584\\[6pt]
\midrule
2&30&0.707&0.6577&0.6956&0.6761&0.7021&0.7042&0.687&0.6866&0.7069&0.6487&0.6997&0.6747\\[6pt]
\midrule
2&50&0.7099&0.6545&0.7029&0.679&0.7006&0.7043&0.6893&0.6838&0.7067&0.6527&0.7019&0.6749\\[6pt]
\midrule
2&100&0.7132&0.6524&0.6947&\textbf{0.698}&0.7024&0.6862&0.6954&0.6996&0.7094&0.6577&0.703&0.6788\\[6pt]
\midrule
2&200&0.7069&0.6456&0.6937&0.6889&0.6888&0.6831&0.693&0.6963&0.7074&0.6602&0.698&0.6748\\[6pt]
\midrule
5&10&0.7485&0.6639&0.7516&0.6681&0.7387&0.7115&0.734&0.7049&0.7529&0.67&0.7451&0.6837\\
\bottomrule
\end{tabular}
\end{table}

\begin{table}[ht!]
\centering
\caption{5-fold Cross Validation Random Forest Model Results: 2 Mins Before Eating}
\label{tbl:Random Forest_Eating_cv_results_min2}
\begin{tabular}{|c|c|c|c|c|c|c|c|c|c|c|c|c|c|}
\toprule
\multirow{2}{*}{Tree Depth} & \multirow{2}{*}{Num. Tree} &\multicolumn{2}{c|}{fold1} & \multicolumn{2}{c|}{fold2} & \multicolumn{2}{c|}{fold3} & \multicolumn{2}{c|}{fold4} & \multicolumn{2}{c|}{fold5}  & \multicolumn{2}{c|}{average} \\\cline{3-14}
&&\multicolumn{1}{c|}{Train} & \multicolumn{1}{c|}{Valid.} & \multicolumn{1}{c|}{Train} & \multicolumn{1}{c|}{Valid.} & \multicolumn{1}{c|}{Train}  & \multicolumn{1}{c|}{Valid.} &\multicolumn{1}{c|}{Train} & \multicolumn{1}{c|}{Valid.} & \multicolumn{1}{c|}{Train} & \multicolumn{1}{c|}{Valid.} & \multicolumn{1}{c|}{Train}  & \multicolumn{1}{c|}{Valid.}\\        \midrule
2&10&0.6709&0.6418&0.6894&0.6785&0.6669&0.6412&0.6565&0.6745&0.6856&0.6549&0.6739&0.6582\\[6pt]
\midrule
2&30&0.7045&0.6551&0.6915&0.6749&0.7029&0.7046&0.6871&0.6868&0.7052&0.6475&0.6982&0.6738\\[6pt]
\midrule
2&50&0.7062&0.6522&0.7004&0.6804&0.6973&0.707&0.6883&0.682&0.7058&0.6493&0.6996&0.6742\\[6pt]
\midrule
2&100&0.7123&0.655&0.696&\textbf{0.6976}&0.695&0.6841&0.7&0.6879&0.7087&0.6554&0.7024&0.676\\[6pt]
\midrule
2&200&0.7061&0.6453&0.6935&0.6933&0.6911&0.6839&0.6924&0.6988&0.7086&0.6559&0.6983&0.6755\\[6pt]
\midrule
5&10&0.7466&0.6617&0.7359&0.6837&0.7365&0.7238&0.7329&0.7073&0.7423&0.6621&0.7389&\textbf{0.6877}\\
\bottomrule
\end{tabular}
\end{table}

\begin{table}[ht!]
\centering
\caption{5-fold Cross Validation Random Forest Model Results: 3 Mins Before Eating}
\label{tbl:Random Forest_Eating_cv_results_min3}
\begin{tabular}{|c|c|c|c|c|c|c|c|c|c|c|c|c|c|}
\toprule
\multirow{2}{*}{Tree Depth} & \multirow{2}{*}{Num. Tree} &\multicolumn{2}{c|}{fold1} & \multicolumn{2}{c|}{fold2} & \multicolumn{2}{c|}{fold3} & \multicolumn{2}{c|}{fold4} & \multicolumn{2}{c|}{fold5}  & \multicolumn{2}{c|}{average} \\\cline{3-14}
&&\multicolumn{1}{c|}{Train} & \multicolumn{1}{c|}{Valid.} & \multicolumn{1}{c|}{Train} & \multicolumn{1}{c|}{Valid.} & \multicolumn{1}{c|}{Train}  & \multicolumn{1}{c|}{Valid.} &\multicolumn{1}{c|}{Train} & \multicolumn{1}{c|}{Valid.} & \multicolumn{1}{c|}{Train} & \multicolumn{1}{c|}{Valid.} & \multicolumn{1}{c|}{Train}  & \multicolumn{1}{c|}{Valid.}\\        \midrule
2&10&0.6877&0.6395&0.6887&0.6812&0.6675&0.6409&0.6812&0.6744&0.7037&0.6537&0.6858&0.6579\\[6pt]
\midrule
2&30&0.7052&0.6544&0.7006&0.6782&0.7002&0.7065&0.6885&0.6871&0.7037&0.6504&0.6996&0.6753\\[6pt]
\midrule
2&50&0.7063&0.653&0.7025&0.6824&0.7&0.7072&0.6909&0.6828&0.7059&0.6515&0.7011&0.6754\\[6pt]
\midrule
2&100&0.7153&0.6489&0.699&0.6917&0.6972&0.6865&0.6976&0.6817&0.7107&0.6581&0.7039&0.6734\\[6pt]
\midrule
2&200&0.7085&0.6454&0.6982&\textbf{0.6939}&0.6885&0.6779&0.6933&0.6921&0.7107&0.6561&0.6998&0.6731\\[6pt]
\midrule
5&10&0.7311&0.6647&0.7411&0.6741&0.7166&0.7141&0.7259&0.6979&0.7444&0.6621&0.7318&0.6826\\
\bottomrule
\end{tabular}
\end{table}

\begin{table}[ht!]
\centering
\caption{5-fold Cross Validation Random Forest Model Results: 4 Mins Before Eating}
\label{tbl:Random Forest_Eating_cv_results_min4}
\begin{tabular}{|c|c|c|c|c|c|c|c|c|c|c|c|c|c|}
\toprule
\multirow{2}{*}{Tree Depth} & \multirow{2}{*}{Num. Tree} &\multicolumn{2}{c|}{fold1} & \multicolumn{2}{c|}{fold2} & \multicolumn{2}{c|}{fold3} & \multicolumn{2}{c|}{fold4} & \multicolumn{2}{c|}{fold5}  & \multicolumn{2}{c|}{average} \\\cline{3-14}
&&\multicolumn{1}{c|}{Train} & \multicolumn{1}{c|}{Valid.} & \multicolumn{1}{c|}{Train} & \multicolumn{1}{c|}{Valid.} & \multicolumn{1}{c|}{Train}  & \multicolumn{1}{c|}{Valid.} &\multicolumn{1}{c|}{Train} & \multicolumn{1}{c|}{Valid.} & \multicolumn{1}{c|}{Train} & \multicolumn{1}{c|}{Valid.} & \multicolumn{1}{c|}{Train}  & \multicolumn{1}{c|}{Valid.}\\        \midrule
2&10&0.6676&0.6352&0.6885&0.6822&0.6653&0.6358&0.6552&0.6881&0.6978&0.6485&0.6749&0.658\\[6pt]
\midrule
2&30&0.7072&0.6536&0.6985&0.6784&0.6882&0.704&0.6873&0.6854&0.7051&0.6502&0.6973&0.6743\\[6pt]
\midrule
2&50&0.7082&0.6507&0.6963&0.6742&0.7032&0.7043&0.6897&0.6802&0.7058&0.649&0.7006&0.6717\\[6pt]
\midrule
2&100&0.7165&0.6548&0.6951&0.6879&0.6971&0.6896&0.6981&0.6802&0.7095&0.6564&0.7033&0.6738\\[6pt]
\midrule
2&200&0.7093&0.6464&0.6956&\textbf{0.692}&0.6973&0.6953&0.6912&0.693&0.7096&0.6605&0.7006&0.6774\\[6pt]
\midrule
5&10&0.7476&0.6579&0.7348&0.6782&0.7144&0.7048&0.7347&0.6994&0.7556&\textbf{0.677}&0.7374&0.6834\\
\bottomrule
\end{tabular}
\end{table}

\begin{table}[ht!]
\centering
\caption{5-fold Cross Validation Random Forest Model Results: 0 Min Before Food Purchasing}
\label{tbl:Random Forest_Food Purchasing_cv_results_min0}
\begin{tabular}{|c|c|c|c|c|c|c|c|c|c|c|c|c|c|}
\toprule
\multirow{2}{*}{Tree Depth} & \multirow{2}{*}{Num. Tree} &\multicolumn{2}{c|}{fold1} & \multicolumn{2}{c|}{fold2} & \multicolumn{2}{c|}{fold3} & \multicolumn{2}{c|}{fold4} & \multicolumn{2}{c|}{fold5}  & \multicolumn{2}{c|}{average} \\\cline{3-14}
&&\multicolumn{1}{c|}{Train} & \multicolumn{1}{c|}{Valid.} & \multicolumn{1}{c|}{Train} & \multicolumn{1}{c|}{Valid.} & \multicolumn{1}{c|}{Train}  & \multicolumn{1}{c|}{Valid.} &\multicolumn{1}{c|}{Train} & \multicolumn{1}{c|}{Valid.} & \multicolumn{1}{c|}{Train} & \multicolumn{1}{c|}{Valid.} & \multicolumn{1}{c|}{Train}  & \multicolumn{1}{c|}{Valid.}\\        \midrule
2&10&0.7314&0.5997&0.7593&0.6818&0.7759&0.653&0.7643&0.6202&0.722&\textbf{0.6988}&0.7506&0.6507\\[6pt]
\midrule
2&30&0.7438&0.589&0.7619&0.7015&0.7967&0.6693&0.793&0.6266&0.7523&0.6554&0.7696&0.6484\\[6pt]
\midrule
2&50&0.7603&0.5789&0.7698&0.7163&0.7967&0.6778&0.7971&0.6161&0.7593&0.6664&0.7767&0.6511\\[6pt]
\midrule
2&100&0.7686&0.551&0.7725&0.7161&0.8008&0.6814&0.793&0.6594&0.764&0.677&0.7798&0.657\\[6pt]
\midrule
2&200&0.781&0.5975&0.7804&0.7123&0.7842&0.6924&0.7951&\textbf{0.6748}&0.757&0.6927&0.7795&\textbf{0.674}\\[6pt]
\midrule
5&10&0.8409&\textbf{0.6176}&0.873&0.6814&0.8693&0.6693&0.8873&0.5942&0.8668&0.6866&0.8675&0.6498\\
\bottomrule
\end{tabular}
\end{table}

\begin{table}[ht!]
\centering
\caption{5-fold Cross Validation Random Forest Model Results: 1 Min Before Food Purchasing}
\label{tbl:Random Forest_Food Purchasing_cv_results_min1}
\begin{tabular}{|c|c|c|c|c|c|c|c|c|c|c|c|c|c|}
\toprule
\multirow{2}{*}{Tree Depth} & \multirow{2}{*}{Num. Tree} &\multicolumn{2}{c|}{fold1} & \multicolumn{2}{c|}{fold2} & \multicolumn{2}{c|}{fold3} & \multicolumn{2}{c|}{fold4} & \multicolumn{2}{c|}{fold5}  & \multicolumn{2}{c|}{average} \\\cline{3-14}
&&\multicolumn{1}{c|}{Train} & \multicolumn{1}{c|}{Valid.} & \multicolumn{1}{c|}{Train} & \multicolumn{1}{c|}{Valid.} & \multicolumn{1}{c|}{Train}  & \multicolumn{1}{c|}{Valid.} &\multicolumn{1}{c|}{Train} & \multicolumn{1}{c|}{Valid.} & \multicolumn{1}{c|}{Train} & \multicolumn{1}{c|}{Valid.} & \multicolumn{1}{c|}{Train}  & \multicolumn{1}{c|}{Valid.}\\        \midrule
2&10&0.7417&0.5824&0.7619&0.6789&0.7656&0.6356&0.7664&0.5997&0.7313&0.6815&0.7534&0.6356\\[6pt]
\midrule
2&30&0.7397&0.5866&0.7619&0.703&0.7967&\textbf{0.711}&0.793&0.5974&0.75&0.691&0.7683&0.6578\\[6pt]
\midrule
2&50&0.7521&0.5582&0.7725&0.7147&0.7988&0.68&0.8094&0.6005&0.764&0.6734&0.7794&0.6454\\[6pt]
\midrule
2&100&0.7769&0.5344&0.7672&0.7281&0.8008&0.6938&0.7807&\textbf{0.655}&0.7593&0.6929&0.777&0.6608\\[6pt]
\midrule
2&200&0.7686&0.5544&0.7619&0.7215&0.7905&0.6894&0.7787&0.655&0.7593&\textbf{0.7014}&0.7718&\textbf{0.6643}\\[6pt]
\midrule
5&10&0.8512&0.6235&0.8677&0.6865&0.8797&0.6365&0.8689&0.5991&0.8551&0.6528&0.8645&0.6397\\
\bottomrule
\end{tabular}
\end{table}

\begin{table}[ht!]
\centering
\caption{5-fold Cross Validation Random Forest Model Results: 2 Mins Before Food Purchasing}
\label{tbl:Random Forest_Food Purchasing_cv_results_min2}
\begin{tabular}{|c|c|c|c|c|c|c|c|c|c|c|c|c|c|}
\toprule
\multirow{2}{*}{Tree Depth} & \multirow{2}{*}{Num. Tree} &\multicolumn{2}{c|}{fold1} & \multicolumn{2}{c|}{fold2} & \multicolumn{2}{c|}{fold3} & \multicolumn{2}{c|}{fold4} & \multicolumn{2}{c|}{fold5}  & \multicolumn{2}{c|}{average} \\\cline{3-14}
&&\multicolumn{1}{c|}{Train} & \multicolumn{1}{c|}{Valid.} & \multicolumn{1}{c|}{Train} & \multicolumn{1}{c|}{Valid.} & \multicolumn{1}{c|}{Train}  & \multicolumn{1}{c|}{Valid.} &\multicolumn{1}{c|}{Train} & \multicolumn{1}{c|}{Valid.} & \multicolumn{1}{c|}{Train} & \multicolumn{1}{c|}{Valid.} & \multicolumn{1}{c|}{Train}  & \multicolumn{1}{c|}{Valid.}\\        \midrule
2&10&0.7583&0.5833&0.7619&0.6627&0.7614&0.6494&0.7787&0.5929&0.7173&0.6948&0.7555&0.6366\\[6pt]
\midrule
2&30&0.7252&0.5751&0.7646&0.7076&0.7884&0.6798&0.7889&0.624&0.757&0.6765&0.7648&0.6526\\[6pt]
\midrule
2&50&0.7438&0.5864&0.7672&0.7212&0.7905&0.6639&0.7951&0.6279&0.7617&0.6693&0.7716&0.6538\\[6pt]
\midrule
2&100&0.7707&0.5563&0.7804&0.7245&0.7967&0.6806&0.7889&0.6707&0.757&0.6666&0.7787&0.6597\\[6pt]
\midrule
2&200&0.7707&0.5587&0.7698&0.7145&0.7905&\textbf{0.6994}&0.7766&0.6671&0.75&0.6726&0.7715&0.6625\\[6pt]
\midrule
5&10&0.8719&0.5815&0.8942&0.6867&0.8734&0.6516&0.875&0.6571&0.8411&\textbf{0.7091}&0.8711&0.6572\\
\bottomrule
\end{tabular}
\end{table}

\begin{table}[ht!]
\centering
\caption{5-fold Cross Validation Random Forest Model Results: 3 Mins Before Food Purchasing}
\label{tbl:Random Forest_Food Purchasing_cv_results_min3}
\begin{tabular}{|c|c|c|c|c|c|c|c|c|c|c|c|c|c|}
\toprule
\multirow{2}{*}{Tree Depth} & \multirow{2}{*}{Num. Tree} &\multicolumn{2}{c|}{fold1} & \multicolumn{2}{c|}{fold2} & \multicolumn{2}{c|}{fold3} & \multicolumn{2}{c|}{fold4} & \multicolumn{2}{c|}{fold5}  & \multicolumn{2}{c|}{average} \\\cline{3-14}
&&\multicolumn{1}{c|}{Train} & \multicolumn{1}{c|}{Valid.} & \multicolumn{1}{c|}{Train} & \multicolumn{1}{c|}{Valid.} & \multicolumn{1}{c|}{Train}  & \multicolumn{1}{c|}{Valid.} &\multicolumn{1}{c|}{Train} & \multicolumn{1}{c|}{Valid.} & \multicolumn{1}{c|}{Train} & \multicolumn{1}{c|}{Valid.} & \multicolumn{1}{c|}{Train}  & \multicolumn{1}{c|}{Valid.}\\        \midrule
2&10&0.7376&0.5288&0.7513&0.6547&0.7614&0.6486&0.7725&0.6215&0.7056&\textbf{0.7104}&0.7457&0.6328\\[6pt]
\midrule
2&30&0.7562&\textbf{0.6144}&0.7619&0.705&0.7946&0.6785&0.7746&0.6276&0.7407&0.6971&0.7656&0.6645\\[6pt]
\midrule
2&50&0.7583&0.5605&0.7672&0.7174&0.8091&0.6848&0.7889&0.6608&0.7547&0.6694&0.7756&0.6586\\[6pt]
\midrule
2&100&0.7769&0.5573&0.7804&0.723&0.7967&0.6842&0.7807&0.6698&0.75&0.6739&0.7769&0.6616\\[6pt]
\midrule
2&200&0.7748&0.5537&0.7725&0.714&0.7946&\textbf{0.696}&0.7787&\textbf{0.6766}&0.7547&0.6686&0.775&0.6618\\[6pt]
\midrule
5&10&0.8636&0.6011&0.8624&0.6919&0.8838&0.6843&0.8914&0.6665&0.8364&0.7094&0.8675&\textbf{0.6706}\\
\bottomrule
\end{tabular}
\end{table}

\begin{table}[ht!]
\centering
\caption{5-fold Cross Validation Random Forest Model Results: 4 Mins Before Food Purchasing}
\label{tbl:Random Forest_Food Purchasing_cv_results_min4}
\begin{tabular}{|c|c|c|c|c|c|c|c|c|c|c|c|c|c|}
\toprule
\multirow{2}{*}{Tree Depth} & \multirow{2}{*}{Num. Tree} &\multicolumn{2}{c|}{fold1} & \multicolumn{2}{c|}{fold2} & \multicolumn{2}{c|}{fold3} & \multicolumn{2}{c|}{fold4} & \multicolumn{2}{c|}{fold5}  & \multicolumn{2}{c|}{average} \\\cline{3-14}
&&\multicolumn{1}{c|}{Train} & \multicolumn{1}{c|}{Valid.} & \multicolumn{1}{c|}{Train} & \multicolumn{1}{c|}{Valid.} & \multicolumn{1}{c|}{Train}  & \multicolumn{1}{c|}{Valid.} &\multicolumn{1}{c|}{Train} & \multicolumn{1}{c|}{Valid.} & \multicolumn{1}{c|}{Train} & \multicolumn{1}{c|}{Valid.} & \multicolumn{1}{c|}{Train}  & \multicolumn{1}{c|}{Valid.}\\        \midrule
2&10&0.7211&\textbf{0.6649}&0.7487&0.644&0.7614&0.6475&0.752&0.6032&0.6986&0.6904&0.7364&0.65\\[6pt]
\midrule
2&30&0.7562&0.5895&0.7646&0.7&0.7863&0.6864&0.7807&0.6248&0.743&0.7144&0.7662&0.663\\[6pt]
\midrule
2&50&0.7562&0.559&0.7646&0.7099&0.7967&0.689&0.7951&0.6471&0.757&0.7008&0.7739&0.6612\\[6pt]
\midrule
2&100&0.7769&0.5594&0.7698&0.7213&0.7905&0.687&0.7807&0.6573&0.7617&0.7354&0.7759&0.6721\\[6pt]
\midrule
2&200&0.7686&0.5662&0.7778&0.7113&0.7925&\textbf{0.6922}&0.7725&\textbf{0.677}&0.7547&\textbf{0.7379}&0.7732&\textbf{0.6769}\\[6pt]
\midrule
5&10&0.8492&0.65&0.8677&0.6717&0.8797&0.6427&0.8996&0.6429&0.8364&0.6962&0.8665&0.6607\\
\bottomrule
\end{tabular}
\end{table}

\begin{table}[ht!]
\centering
\caption{5-fold Cross Validation Gradient Boosting Model Results: 0 Min Before Eating}
\label{tbl:Gradient Boosting_Eating_cv_results_min0}
\begin{tabular}{|c|c|c|c|c|c|c|c|c|c|c|c|c|c|}
\toprule
\multirow{2}{*}{Tree Depth} & \multirow{2}{*}{Num. Tree} &\multicolumn{2}{c|}{fold1} & \multicolumn{2}{c|}{fold2} & \multicolumn{2}{c|}{fold3} & \multicolumn{2}{c|}{fold4} & \multicolumn{2}{c|}{fold5}  & \multicolumn{2}{c|}{average} \\\cline{3-14}
&&\multicolumn{1}{c|}{Train} & \multicolumn{1}{c|}{Valid.} & \multicolumn{1}{c|}{Train} & \multicolumn{1}{c|}{Valid.} & \multicolumn{1}{c|}{Train}  & \multicolumn{1}{c|}{Valid.} &\multicolumn{1}{c|}{Train} & \multicolumn{1}{c|}{Valid.} & \multicolumn{1}{c|}{Train} & \multicolumn{1}{c|}{Valid.} & \multicolumn{1}{c|}{Train}  & \multicolumn{1}{c|}{Valid.}\\        \midrule
1&10&0.6865&0.646&0.6513&0.6566&0.6619&0.648&0.6744&0.668&0.6858&0.6442&0.672&0.6525\\[6pt]
\midrule
1&30&0.7192&0.6732&0.7015&0.6784&0.6934&0.6966&0.6927&0.6902&0.7138&0.656&0.7041&0.6789\\[6pt]
\midrule
1&50&0.7289&0.6761&0.7109&0.6878&0.7025&0.7141&0.7148&0.6991&0.7322&0.6698&0.7178&0.6894\\[6pt]
\midrule
1&100&0.739&0.678&0.7227&0.6855&0.7233&0.7193&0.7303&0.713&0.7391&0.6723&0.7309&0.6936\\[6pt]
\midrule
1&200&0.748&\textbf{0.6888}&0.7432&\textbf{0.7032}&0.7315&0.728&0.7353&0.724&0.7436&0.6925&0.7403&\textbf{0.7073}\\[6pt]
\midrule
2&10&0.7068&0.6656&0.7041&0.685&0.695&0.7123&0.69&0.6982&0.7035&0.6742&0.6999&0.6871\\
\bottomrule
\end{tabular}
\end{table}

\begin{table}[ht!]
\centering
\caption{5-fold Cross Validation Gradient Boosting Model Results: 1 Min Before Eating}
\label{tbl:Gradient Boosting_Eating_cv_results_min1}
\begin{tabular}{|c|c|c|c|c|c|c|c|c|c|c|c|c|c|}
\toprule
\multirow{2}{*}{Tree Depth} & \multirow{2}{*}{Num. Tree} &\multicolumn{2}{c|}{fold1} & \multicolumn{2}{c|}{fold2} & \multicolumn{2}{c|}{fold3} & \multicolumn{2}{c|}{fold4} & \multicolumn{2}{c|}{fold5}  & \multicolumn{2}{c|}{average} \\\cline{3-14}
&&\multicolumn{1}{c|}{Train} & \multicolumn{1}{c|}{Valid.} & \multicolumn{1}{c|}{Train} & \multicolumn{1}{c|}{Valid.} & \multicolumn{1}{c|}{Train}  & \multicolumn{1}{c|}{Valid.} &\multicolumn{1}{c|}{Train} & \multicolumn{1}{c|}{Valid.} & \multicolumn{1}{c|}{Train} & \multicolumn{1}{c|}{Valid.} & \multicolumn{1}{c|}{Train}  & \multicolumn{1}{c|}{Valid.}\\        \midrule
1&10&0.6877&0.645&0.6509&0.6572&0.6615&0.6479&0.6754&0.6675&0.687&0.6444&0.6725&0.6524\\[6pt]
\midrule
1&30&0.7192&0.6732&0.7019&0.6801&0.6933&0.6958&0.6941&0.69&0.7154&0.6576&0.7048&0.6793\\[6pt]
\midrule
1&50&0.7279&0.6778&0.7107&0.6874&0.703&0.7156&0.7125&0.696&0.7317&0.6688&0.7171&0.6891\\[6pt]
\midrule
1&100&0.7384&0.6779&0.7232&0.686&0.7197&0.7192&0.7304&0.7083&0.7392&0.6802&0.7302&0.6943\\[6pt]
\midrule
1&200&0.7491&0.6869&0.7442&\textbf{0.7031}&0.7295&0.7282&0.7334&0.7221&0.7442&\textbf{0.6878}&0.7401&\textbf{0.7056}\\[6pt]
\midrule
2&10&0.7062&0.6647&0.7041&0.6852&0.695&0.7106&0.6898&0.6978&0.7037&0.6749&0.6998&0.6866\\
\bottomrule
\end{tabular}
\end{table}

\begin{table}[ht!]
\centering
\caption{5-fold Cross Validation Gradient Boosting Model Results: 2 Mins Before Eating}
\label{tbl:Gradient Boosting_Eating_cv_results_min2}
\begin{tabular}{|c|c|c|c|c|c|c|c|c|c|c|c|c|c|}
\toprule
\multirow{2}{*}{Tree Depth} & \multirow{2}{*}{Num. Tree} &\multicolumn{2}{c|}{fold1} & \multicolumn{2}{c|}{fold2} & \multicolumn{2}{c|}{fold3} & \multicolumn{2}{c|}{fold4} & \multicolumn{2}{c|}{fold5}  & \multicolumn{2}{c|}{average} \\\cline{3-14}
&&\multicolumn{1}{c|}{Train} & \multicolumn{1}{c|}{Valid.} & \multicolumn{1}{c|}{Train} & \multicolumn{1}{c|}{Valid.} & \multicolumn{1}{c|}{Train}  & \multicolumn{1}{c|}{Valid.} &\multicolumn{1}{c|}{Train} & \multicolumn{1}{c|}{Valid.} & \multicolumn{1}{c|}{Train} & \multicolumn{1}{c|}{Valid.} & \multicolumn{1}{c|}{Train}  & \multicolumn{1}{c|}{Valid.}\\        \midrule
1&10&0.6886&0.6443&0.6504&0.6575&0.6612&0.6479&0.6761&0.667&0.6876&0.6444&0.6728&0.6522\\[6pt]
\midrule
1&30&0.719&0.6734&0.7026&0.6798&0.6913&0.6921&0.6944&0.6885&0.7159&0.6575&0.7046&0.6783\\[6pt]
\midrule
1&50&0.7264&0.6767&0.7076&0.6851&0.702&0.7138&0.7137&0.6952&0.7311&0.668&0.7162&0.6877\\[6pt]
\midrule
1&100&0.7415&0.6796&0.7198&0.6862&0.7201&0.72&0.7269&0.709&0.7351&0.6695&0.7287&0.6929\\[6pt]
\midrule
1&200&0.7461&0.6857&0.7459&\textbf{0.6971}&0.7309&0.7322&0.736&\textbf{0.7207}&0.7459&0.6851&0.741&\textbf{0.7042}\\[6pt]
\midrule
2&10&0.7081&0.667&0.7006&0.6868&0.6949&0.7105&0.6898&0.6977&0.7036&0.675&0.6994&0.6874\\
\bottomrule
\end{tabular}
\end{table}

\begin{table}[ht!]
\centering
\caption{5-fold Cross Validation Gradient Boosting Model Results: 3 Mins Before Eating}
\label{tbl:Gradient Boosting_Eating_cv_results_min3}
\begin{tabular}{|c|c|c|c|c|c|c|c|c|c|c|c|c|c|}
\toprule
\multirow{2}{*}{Tree Depth} & \multirow{2}{*}{Num. Tree} &\multicolumn{2}{c|}{fold1} & \multicolumn{2}{c|}{fold2} & \multicolumn{2}{c|}{fold3} & \multicolumn{2}{c|}{fold4} & \multicolumn{2}{c|}{fold5}  & \multicolumn{2}{c|}{average} \\\cline{3-14}
&&\multicolumn{1}{c|}{Train} & \multicolumn{1}{c|}{Valid.} & \multicolumn{1}{c|}{Train} & \multicolumn{1}{c|}{Valid.} & \multicolumn{1}{c|}{Train}  & \multicolumn{1}{c|}{Valid.} &\multicolumn{1}{c|}{Train} & \multicolumn{1}{c|}{Valid.} & \multicolumn{1}{c|}{Train} & \multicolumn{1}{c|}{Valid.} & \multicolumn{1}{c|}{Train}  & \multicolumn{1}{c|}{Valid.}\\        \midrule
1&10&0.6901&0.6433&0.6818&0.6601&0.6608&0.6477&0.6775&0.6667&0.6885&0.6449&0.6797&0.6525\\[6pt]
\midrule
1&30&0.7186&0.6733&0.7026&0.6805&0.6916&0.6911&0.6948&0.6874&0.7161&0.6563&0.7047&0.6777\\[6pt]
\midrule
1&50&0.7243&0.6774&0.7078&0.6853&0.7024&0.71&0.7171&0.6956&0.7163&0.6585&0.7136&0.6853\\[6pt]
\midrule
1&100&0.7396&0.6781&0.7198&0.6858&0.7207&0.7171&0.7284&0.7052&0.7389&0.6699&0.7295&0.6912\\[6pt]
\midrule
1&200&0.7409&0.6818&0.7452&\textbf{0.6965}&0.7298&0.7288&0.7345&\textbf{0.7204}&0.7465&0.6815&0.7394&\textbf{0.7018}\\[6pt]
\midrule
2&10&0.7163&0.6721&0.7007&0.6871&0.6952&0.711&0.6903&0.6974&0.7143&0.6802&0.7033&0.6896\\
\bottomrule
\end{tabular}
\end{table}

\begin{table}[ht!]
\centering
\caption{5-fold Cross Validation Gradient Boosting Model Results: 4 Mins Before Eating}
\label{tbl:Gradient Boosting_Eating_cv_results_min4}
\begin{tabular}{|c|c|c|c|c|c|c|c|c|c|c|c|c|c|}
\toprule
\multirow{2}{*}{Tree Depth} & \multirow{2}{*}{Num. Tree} &\multicolumn{2}{c|}{fold1} & \multicolumn{2}{c|}{fold2} & \multicolumn{2}{c|}{fold3} & \multicolumn{2}{c|}{fold4} & \multicolumn{2}{c|}{fold5}  & \multicolumn{2}{c|}{average} \\\cline{3-14}
&&\multicolumn{1}{c|}{Train} & \multicolumn{1}{c|}{Valid.} & \multicolumn{1}{c|}{Train} & \multicolumn{1}{c|}{Valid.} & \multicolumn{1}{c|}{Train}  & \multicolumn{1}{c|}{Valid.} &\multicolumn{1}{c|}{Train} & \multicolumn{1}{c|}{Valid.} & \multicolumn{1}{c|}{Train} & \multicolumn{1}{c|}{Valid.} & \multicolumn{1}{c|}{Train}  & \multicolumn{1}{c|}{Valid.}\\        \midrule
1&10&0.6912&0.644&0.6792&0.6569&0.6605&0.6477&0.679&0.6662&0.6896&0.6453&0.6799&0.652\\[6pt]
\midrule
1&30&0.7168&0.6724&0.703&0.6804&0.6914&0.6905&0.6942&0.687&0.7165&0.6566&0.7044&0.6774\\[6pt]
\midrule
1&50&0.7264&0.6617&0.7109&0.685&0.7004&0.7075&0.7142&0.6918&0.7284&0.6649&0.7161&0.6822\\[6pt]
\midrule
1&100&0.7355&0.6767&0.7204&0.6853&0.7167&0.7161&0.7266&0.7035&0.7373&0.6801&0.7273&0.6923\\[6pt]
\midrule
1&200&0.7392&0.6814&0.745&\textbf{0.6955}&0.7289&0.7287&0.7322&0.7173&0.7446&0.6804&0.738&\textbf{0.7007}\\[6pt]
\midrule
2&10&0.7166&0.6714&0.701&0.6873&0.697&0.7115&0.6957&0.6983&0.7039&0.6767&0.7028&0.689\\
\bottomrule
\end{tabular}
\end{table}

\begin{table}[ht!]
\centering
\caption{5-fold Cross Validation Gradient Boosting Model Results: 0 Min Before Food Purchasing}
\label{tbl:Gradient Boosting_Food Purchasing_cv_results_min0}
\begin{tabular}{|c|c|c|c|c|c|c|c|c|c|c|c|c|c|}
\toprule
\multirow{2}{*}{Tree Depth} & \multirow{2}{*}{Num. Tree} &\multicolumn{2}{c|}{fold1} & \multicolumn{2}{c|}{fold2} & \multicolumn{2}{c|}{fold3} & \multicolumn{2}{c|}{fold4} & \multicolumn{2}{c|}{fold5}  & \multicolumn{2}{c|}{average} \\\cline{3-14}
&&\multicolumn{1}{c|}{Train} & \multicolumn{1}{c|}{Valid.} & \multicolumn{1}{c|}{Train} & \multicolumn{1}{c|}{Valid.} & \multicolumn{1}{c|}{Train}  & \multicolumn{1}{c|}{Valid.} &\multicolumn{1}{c|}{Train} & \multicolumn{1}{c|}{Valid.} & \multicolumn{1}{c|}{Train} & \multicolumn{1}{c|}{Valid.} & \multicolumn{1}{c|}{Train}  & \multicolumn{1}{c|}{Valid.}\\        \midrule
1&10&0.6818&0.5999&0.6984&0.6709&0.6909&0.6474&0.7029&0.6241&0.7243&0.6466&0.6997&0.6378\\[6pt]
\midrule
1&30&0.7376&0.5819&0.7487&0.7198&0.7822&0.6863&0.7541&0.6611&0.7453&0.6601&0.7536&0.6618\\[6pt]
\midrule
1&50&0.7624&0.6253&0.7725&0.7249&0.7905&0.6281&0.7602&0.6747&0.764&0.6766&0.7699&0.6659\\[6pt]
\midrule
1&100&0.7955&0.571&0.7989&0.736&0.8237&0.6028&0.8074&0.6702&0.7944&0.6692&0.804&0.6498\\[6pt]
\midrule
1&200&0.8306&0.5739&0.8492&\textbf{0.7493}&0.8506&0.5818&0.8504&0.6801&0.8364&0.6612&0.8435&0.6493\\[6pt]
\midrule
2&10&0.75&\textbf{0.6747}&0.7831&0.7093&0.7676&\textbf{0.6912}&0.7848&\textbf{0.7011}&0.7407&\textbf{0.7119}&0.7652&\textbf{0.6977}\\
\bottomrule
\end{tabular}
\end{table}

\begin{table}[ht!]
\centering
\caption{5-fold Cross Validation Gradient Boosting Model Results: 1 Min Before Food Purchasing}
\label{tbl:Gradient Boosting_Food Purchasing_cv_results_min1}
\begin{tabular}{|c|c|c|c|c|c|c|c|c|c|c|c|c|c|}
\toprule
\multirow{2}{*}{Tree Depth} & \multirow{2}{*}{Num. Tree} &\multicolumn{2}{c|}{fold1} & \multicolumn{2}{c|}{fold2} & \multicolumn{2}{c|}{fold3} & \multicolumn{2}{c|}{fold4} & \multicolumn{2}{c|}{fold5}  & \multicolumn{2}{c|}{average} \\\cline{3-14}
&&\multicolumn{1}{c|}{Train} & \multicolumn{1}{c|}{Valid.} & \multicolumn{1}{c|}{Train} & \multicolumn{1}{c|}{Valid.} & \multicolumn{1}{c|}{Train}  & \multicolumn{1}{c|}{Valid.} &\multicolumn{1}{c|}{Train} & \multicolumn{1}{c|}{Valid.} & \multicolumn{1}{c|}{Train} & \multicolumn{1}{c|}{Valid.} & \multicolumn{1}{c|}{Train}  & \multicolumn{1}{c|}{Valid.}\\        \midrule
1&10&0.6798&0.5937&0.6984&0.6705&0.6909&0.6474&0.7213&0.6972&0.6799&0.6558&0.6941&0.6529\\[6pt]
\midrule
1&30&0.7335&0.6036&0.7513&0.7198&0.7718&0.6785&0.752&0.6623&0.743&0.6608&0.7503&0.665\\[6pt]
\midrule
1&50&0.7583&0.6244&0.7513&0.728&0.7884&0.6636&0.7643&0.6656&0.7547&0.6967&0.7634&0.6757\\[6pt]
\midrule
1&100&0.7893&0.5754&0.8122&0.7147&0.8216&0.6168&0.8135&0.6192&0.7874&0.6872&0.8048&0.6427\\[6pt]
\midrule
1&200&0.8326&0.5704&0.8413&\textbf{0.7331}&0.8506&0.599&0.8422&0.6436&0.8294&0.6317&0.8392&0.6356\\[6pt]
\midrule
2&10&0.7583&\textbf{0.7206}&0.7751&0.717&0.7759&0.6891&0.7889&\textbf{0.7093}&0.75&\textbf{0.7321}&0.7697&\textbf{0.7136}\\
\bottomrule
\end{tabular}
\end{table}

\begin{table}[ht!]
\centering
\caption{5-fold Cross Validation Gradient Boosting Model Results: 2 Mins Before Food Purchasing}
\label{tbl:Gradient Boosting_Food Purchasing_cv_results_min2}
\begin{tabular}{|c|c|c|c|c|c|c|c|c|c|c|c|c|c|}
\toprule
\multirow{2}{*}{Tree Depth} & \multirow{2}{*}{Num. Tree} &\multicolumn{2}{c|}{fold1} & \multicolumn{2}{c|}{fold2} & \multicolumn{2}{c|}{fold3} & \multicolumn{2}{c|}{fold4} & \multicolumn{2}{c|}{fold5}  & \multicolumn{2}{c|}{average} \\\cline{3-14}
&&\multicolumn{1}{c|}{Train} & \multicolumn{1}{c|}{Valid.} & \multicolumn{1}{c|}{Train} & \multicolumn{1}{c|}{Valid.} & \multicolumn{1}{c|}{Train}  & \multicolumn{1}{c|}{Valid.} &\multicolumn{1}{c|}{Train} & \multicolumn{1}{c|}{Valid.} & \multicolumn{1}{c|}{Train} & \multicolumn{1}{c|}{Valid.} & \multicolumn{1}{c|}{Train}  & \multicolumn{1}{c|}{Valid.}\\        \midrule
1&10&0.6467&\textbf{0.6599}&0.6984&0.6705&0.6909&0.6471&0.7234&0.6965&0.7126&0.6165&0.6944&0.6581\\[6pt]
\midrule
1&30&0.7376&0.5933&0.746&0.7232&0.778&0.6809&0.7541&0.7019&0.75&0.6566&0.7531&0.6712\\[6pt]
\midrule
1&50&0.7521&0.6193&0.7566&0.7252&0.7842&0.6619&0.7664&0.6936&0.7617&0.6851&0.7642&0.677\\[6pt]
\midrule
1&100&0.7934&0.5637&0.8175&0.7358&0.8174&0.6202&0.8053&0.6709&0.7804&0.6672&0.8028&0.6516\\[6pt]
\midrule
1&200&0.8202&0.5778&0.8413&\textbf{0.7437}&0.8548&0.6928&0.8463&0.6677&0.8294&0.6516&0.8384&0.6667\\[6pt]
\midrule
2&10&0.75&0.6369&0.7751&0.7169&0.7635&0.6814&0.7869&\textbf{0.7029}&0.7523&\textbf{0.714}&0.7656&\textbf{0.6904}\\
\bottomrule
\end{tabular}
\end{table}

\begin{table}[ht!]
\centering
\caption{5-fold Cross Validation Gradient Boosting Model Results: 3 Mins Before Food Purchasing}
\label{tbl:Gradient Boosting_Food Purchasing_cv_results_min3}
%
\end{table}

\begin{table}[ht!]
\centering
\caption{5-fold Cross Validation Gradient Boosting Model Results: 4 Mins Before Food Purchasing}
\label{tbl:Gradient Boosting_Food Purchasing_cv_results_min4}
%
\end{table}

\end{document}